\documentclass[letterpaper, 10 pt, journal, twoside]{IEEEtran}

\IEEEoverridecommandlockouts 

\usepackage{graphics} 
\usepackage{mathptmx} 
\usepackage{amsmath} 
\usepackage{amssymb}  
\usepackage[export]{adjustbox} 
\usepackage{hyperref}
\usepackage{cleveref}
\usepackage{mathtools}
\usepackage{bm}
\usepackage{subcaption} 
\usepackage{tikz}

\usepackage{xcolor}
\usepackage{graphics}

\usepackage[font=small]{caption}

\title{RGB-D SLAM in Indoor Planar Environments with Multiple Large Dynamic Objects
}

\author{Ran Long$^{1,*}$, Christian Rauch$^{1}$, Tianwei Zhang$^{2}$, Vladimir Ivan$^{1}$, Tin Lun Lam$^{2,3}$, Sethu Vijayakumar$^{1,*}$%
\thanks{Manuscript received: February, 23, 2022; Revised April, 18, 2022; Accepted June, 11, 2022.}%
\thanks{This paper was recommended for publication by Editor Tamim Asfour upon evaluation of the Associate Editor and Reviewers' comments. This research is supported by the European Union’s Horizon 2020 research and innovation programme under grant agreement No 101017008 (Harmony) and the Alan Turing Institute.}%
\thanks{$^{1}$The authors are with the Institute of Perception, Action and Behaviour, School of Informatics, University of Edinburgh, Edinburgh, EH8 9AB, U.K.}%
\thanks{$^{2}$The authors are with the Shenzhen Institute of Artificial Intelligence and Robotics for Society (AIRS).}%
\thanks{$^{3}$The author is with the School of Science and Engineering, the Chinese University of Hong Kong, Shenzhen.}
\thanks{$^{*}$Corresponding authors: Ran Long (\texttt{Ran.Long@ed.ac.uk}), Sethu Vijayakumar (\texttt{Sethu.Vijayakumar@ed.ac.uk}).}%
\thanks{Digital Object Identifier (DOI): see top of this page.}%
}

\begin{document}

\maketitle

\begin{abstract}
This work presents a novel dense RGB-D SLAM approach for dynamic planar environments that enables simultaneous multi-object tracking, camera localisation and background reconstruction. Previous dynamic SLAM methods either rely on semantic segmentation to directly detect dynamic objects; or assume that dynamic objects occupy a smaller proportion of the camera view than the static background and can, therefore, be removed as outliers. With the aid of camera motion prior, our approach enables dense SLAM when the camera view is largely occluded by multiple dynamic objects. The dynamic planar objects are separated by their different rigid motions and tracked independently. The remaining dynamic non-planar areas are removed as outliers and not mapped into the background. The evaluation demonstrates that our approach outperforms the state-of-the-art methods in terms of localisation, mapping, dynamic segmentation and object tracking. We also demonstrate its robustness to large drift in the camera motion prior.
\end{abstract}


\section{Introduction}
\IEEEPARstart{S}{imultaneous} localisation and mapping (SLAM) is one of the core components in autonomous robots and virtual reality applications. In indoor environments, planes are common man-made features. Planar SLAM methods have used the characteristics of planes to reduce long-term drift and improve the accuracy of localisation \cite{Li2021PlanarSLAM, zhang2019point}. However, these methods assume that the environment is static -- an assumption that is violated when the robot works in conjunction with other humans or robots, or manipulates objects in semi-automated warehouses. 

The core problem of enabling SLAM in dynamic environments while differentiating multiple dynamic objects involves several challenges:
\begin{enumerate}
    \item There are usually an unknown number of third-party motions in addition to the camera motion in dynamic environments. The number of motions or dynamic objects is also changing. 
    
    \item Static background is often assumed to account for the major proportion of the camera view. However, without semantic segmentation, dynamic objects that occupy a large proportion of the camera view can end up being classified as the static background.
    
    \item The majority of the colour and depth information can be occluded by dynamic objects and the remaining static parts of the visual input may not be enough to support accurate camera ego-motion estimation.
\end{enumerate}

\begin{figure}[t]
    \centering
    \setlength{\tabcolsep}{0pt}
    \begin{tabular}{ccc}
    \includegraphics[width=0.33\linewidth,frame,valign=m]{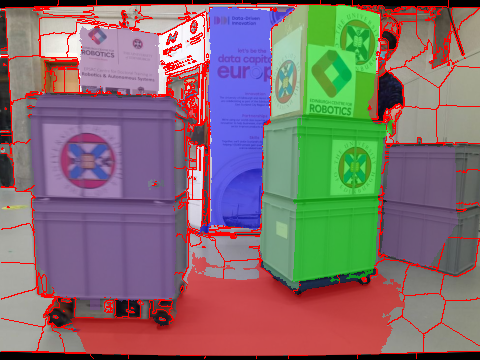} & 
    \includegraphics[width=0.33\linewidth,frame,valign=m]{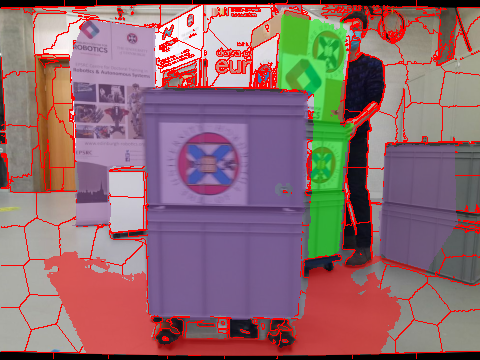} & 
    \includegraphics[width=0.33\linewidth,frame,valign=m]{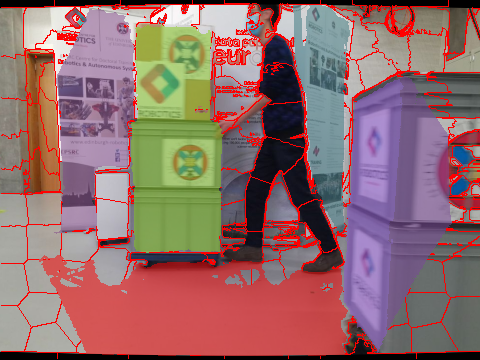}  
    \end{tabular}

    \caption{
    Hierarchical segmentation based on planes and non-planar areas. The planes are extracted from the depth map and the non-planar areas are represented by a set of super-pixels. 
    }
    \label{fig:teaser}
\end{figure}

Many dynamic SLAM methods have considered multiple dynamic objects \cite{huang2020clustervo,judd2018multimotion, runz2017co}, but either rely on semantic segmentation or assume that the static background is the largest rigid body in the camera view. To concurrently solve these problems, we propose a hierarchical representation of images that extracts planes from planar areas and over-segments non-planar areas into super-pixels (\Cref{fig:teaser}). We consequently segment and track multiple dynamic planar rigid objects, and remove dynamic non-planar objects to enable camera localisation and mapping. For this, we assume that planes occupy a major fraction of the environment, including the static background and rigid dynamic objects. In addition, the camera motion can be distinguished from other third-party motions by a tightly-coupled camera motion prior from robot odometry.

In summary, this work contributes:
\begin{enumerate}
    \item a new methodology for online multimotion segmentation based on planes in indoor dynamic environments,
    \item a novel pipeline that simultaneously tracks multiple planar rigid objects, estimates camera ego-motion and reconstructs the static background,
    \item a RGB-D SLAM method that is robust to large-occluded camera view caused by multiple large dynamic objects.
\end{enumerate}

\section{Related Work}

\subsection{Dynamic SLAM}
Dynamic SLAM methods can be categorised into outlier-based, semantic-based, multimotion and proprioception-aided methods.

\textit{Outlier-based} methods assume that the static background occupies the major component in the camera view and dynamic objects can be robustly removed during camera tracking. Joint-VO-SF (JF) \cite{jaimez2017fast} over-segments images into clusters and classifies each cluster as either static or dynamic by comparing the depth and intensity residuals. StaticFusion (SF) \cite{scona2018staticfusion} introduces a continuous score to represent the probability that a cluster is static. The scores are used to segment the static background for camera localisation and mapping, while the remaining dynamic parts are discarded as outliers. Co-Fusion (CF) \cite{runz2017co} and MultiMotionFusion \cite{rauch2022sparse} can further model the outliers as new objects and track them independently. 

\textit{Semantic-based} methods directly detect dynamic objects from the semantic segmentation. Based on Mask R-CNN \cite{he2017mask}, which provides pixel-wise object segmentation, EM-Fusion \cite{Strecke2019EMFusion} integrates object tracking and SLAM into a single expectation maximisation (EM) framework. ClusterVO \cite{huang2020clustervo} can track camera ego-motion and multiple rigidly moving clusters simultaneously by combining semantic bounding boxes and ORB features \cite{rublee2011orb}. DynaSLAM II \cite{bescos2021dynaslam} further integrates the multi-object tracking (MOT) and SLAM into a tightly-coupled formulation to improve its performance on both problems. However, these methods require that the object detector is pre-trained on a dataset which includes these objects or that the object model is provided in advance.

\textit{Multimotion} methods explicitly model the dynamic component as third-party rigid motions and segment dynamic objects by their different motions against the camera motion. Multimotion visual odometry (MVO) \cite{judd2018multimotion} is an online multimotion segmentation and tracking method based on sparse keypoints. It iteratively samples all keypoints using RANSAC to generate motion hypotheses and automatically merge them when the merging can decrease the total energy. In addition to RANSAC, motion hypotheses can be generated from the increased N-points using a grid-based scene flow \cite{lee2019robust}. Instead of merging motion hypotheses, DymSLAM \cite{wang2020dymslam} computes a residual matrix from hypothetical motion models and directly estimates the number of dynamic objects from the residual matrix. The dense segmentation of multiple dynamic objects based on super-pixels \cite{achanta2012slic} is acquired after the number of objects is specified. However, all these methods assume that the largest body of motion in the camera view is the static background and they have not demonstrated their effectiveness when the major proportion of the camera view is occluded by dynamic objects.

Robot proprioception, such as IMU and wheel odometry, can be fused with visual sensors to improve the accuracy and robustness of localisation in dynamic environments \cite{kim2015visual, long2021rigidfusion}. Kim et al. \cite{kim2015visual} uses the camera motion prior from an IMU to compensate for the camera motion and select static keypoints based on motion vectors. RigidFusion (RF) \cite{long2021rigidfusion} uses the camera motion prior from wheel odometry and additional object motion priors to enable SLAM with single dynamic object reconstruction when the major part of the camera view is occluded.
However, these methods are unable to track multiple dynamic objects independently.

\subsection{Planar SLAM}
Planar features have been widely used in indoor dynamic SLAM methods. Infinite planes can be used as landmarks in the pose graph SLAM problem \cite{kaess2015simultaneous}. Based on keyframe management, a global dense planar map can be reconstructed using only a single CPU \cite{hsiao2017keyframe}. Planes can also be combined with keypoints and lines \cite{zhang2019point,Li2021PlanarSLAM} for more robust camera tracking. In structured environments, planes have been demonstrated to significantly reduce accumulated rotational drift under Manhattan world (MW) assumption \cite{Li2021PlanarSLAM}. All these methods assume static environments, because planes in the indoor environment, like walls, are often static. However, this assumption is violated when planar objects, such as boxes, are transported or manipulated by humans or robots.

\section{Methodology}
\subsection{Overview and notation}
\begin{figure*}[t]
    \centering
    \includegraphics[width=\linewidth]{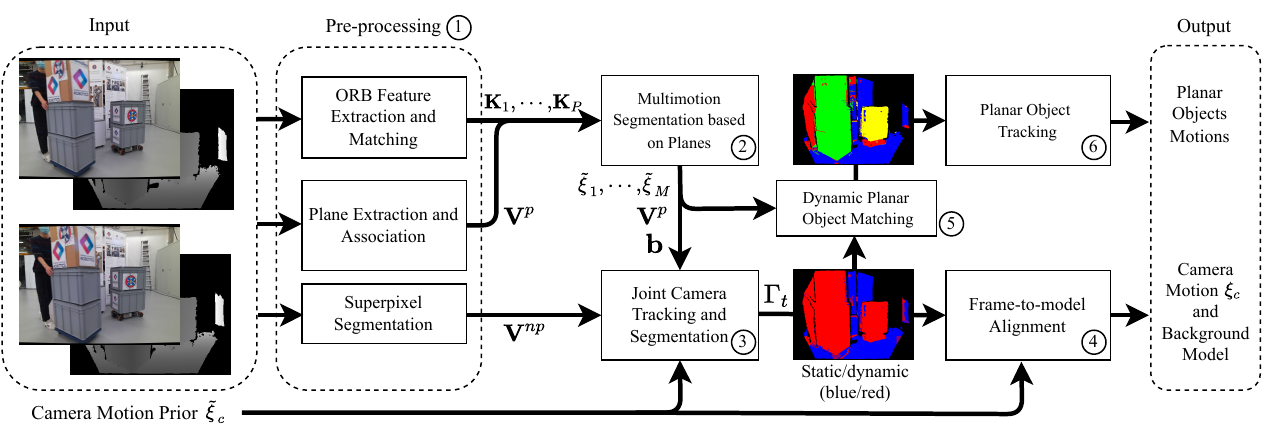}
    \caption{The pipeline of our proposed method. (1) We first represent the input image in the current frame $t$ as a combination of planes and super-pixels. The ORB features \cite{rublee2011orb} are extracted and matched to the previous frame. (2) Planes with similar rigid motions are clustered into $M$ planar rigid bodies and their corresponding egocentric motions are estimated respectively. However, we are uncertain which planar rigid body belongs to the static background. (3) We, therefore, jointly separate the static background from the planes and super-pixels, and estimate the camera motion via frame-to-frame alignment. (4) The static part is used to reconstruct the background and refine the camera motion. (5,6) Dynamic non-planar super-pixels are removed as outliers, while dynamic planar rigid bodies are matched with planar rigid bodies in the previous frame. The matched planar rigid body is tracked using RANSAC on their ORB features and plane parameters.}
    \label{fig:pipeline}
\end{figure*}

The overview of our pipeline is illustrated in \Cref{fig:pipeline}. Our method takes RGB-D image pairs from two consecutive frames. At the $t$-th frame, we have a depth image $D_t$ and an intensity image $I_t$ computed from the colour image.

A plane is represented in the Hessian form $\Pi = (\mathbf{n}^T, d)^T$, where $\mathbf{n} = (n_x, n_y, n_z)$ is the normal of the plane and $d$ is the perpendicular distance between the plane and camera origin. For each image frame, we extract planes directly from the depth map using PEAC \cite{feng2014fast}, which can provide the number of planes $P$, the pixel-wise segmentation of planes $\mathbf{V}^p : \{V^p_i|i\in[1, P]\}$ and their corresponding plane parameters. After plane extraction, the remaining non-planar areas are over-segmented into $S$ super-pixels: $\mathbf{V}^{np} : \{V^{np}_i|i\in[1, S]\}$. 

For the $i$-th plane, we extract a set of keypoints $\mathbf{K}_i$ using ORB features \cite{mur2017orb}. We then conduct multimotion segmentation on planar areas $\mathbf{V}^p$ and cluster the planes into $M$ planar rigid bodies of different motions. For simplicity, we name the camera motion relative to objects' egocentric frames as \textit{object egocentric motion} \cite{judd2018multimotion} and denote them as  $^{ego}\mathbf{T}=\{\tilde{\xi}_m \in \mathfrak{se}(3)| m \in [1,M] \}$. Since the static background may not be the largest rigid body in the camera view, we use the camera motion prior $\tilde{\xi}_c$ with potential drift to simultaneously classify all planes and super-pixels into either static or dynamic, and estimate the camera ego-motion.

We use the score $\gamma_i \in [0,1]$ to represent the probability that a plane or super-pixel is static. The scores $\bm{\gamma}: \{\gamma_i | i \in [1,P+S]\}$ are assigned to each plane and super-pixel, where $\{\gamma_i | i \in [1,P]\}$ and $\{\gamma_i | i \in [P+1,P+S]\}$ represent the scores of planes and super-pixels respectively. At time $t$, the pixel-wise static dynamic segmentation $\Gamma_t \in \mathbb{R}^{w \times h}$ can be estimated from $\bm{\gamma}$. The static parts of intensity and depth images are used to estimate the camera motion $\xi_c \in \mathfrak{se}(3)$. The dynamic planar rigid bodies are used to track dynamic objects. The non-planar dynamic super-pixels, such as humans, are removed as outliers.

The world-, camera-, and the $m$-th object-frames are $W$, $C$ and $O^m$ respectively. The camera motion $T(\xi_c) \coloneqq exp(\xi_c)$ is $T_{C_{t-1} C_t} = T_{WC_{t-1}}^{-1}T_{WC_t}$, which transforms homogeneous coordinates of a point in the current camera frame $C_{t}$ to the previous frame $C_{t-1}$. The function $exp(\xi)$ is the matrix exponential map for Lie group $SE(3)$. The $m$-th object egocentric transformations is the camera motion relative to this object \cite{judd2018multimotion}:
\begin{align}
T(\tilde{\xi}_m) = {^{m}}T^{-1}_{C_{t-1}C_t} = T^{-1}_{O^m_{t-1}C_{t-1}}T_{O^m_tC_t}.
\end{align}

\subsection{Multimotion segmentation based on planes}
To extract planes, we transform the depth map to a point cloud and cluster connected groups of points with close normal directions using the method from \cite{feng2014fast}. To match plane $i$ in the current frame with one in the previous frame, we first estimate the angle and point-to-plane distance between plane $i$ and all planes in the previous frame. A plane is chosen as a candidate if the angle and distance are below 10 degrees and 0.1 m respectively, which is the same as \cite{zhang2019point, Li2021PlanarSLAM}. However, rather than choosing the plane with minimal distance \cite{zhang2019point, Li2021PlanarSLAM}, we further consider overlap proportion between two planes using the Jaccard index, $J(V_1, V_2) = \frac{|V_1 \cap V_2|}{|V_1 \cup V_2|}$, where $|V_1|$ is the number of pixels in planar segment $1$. We choose the candidate plane that has the maximal Jaccard index as the matched plane for plane $i$ and denote it as plane $i'$.

To estimate object egocentric motion $T_i = {^i}T^{-1}_{C_{t-1}C_t}$ of plane $i$, we extract and match ORB keypoints from plane $i$ and $i'$. The error function is defined as: 
\begin{equation}
    e_{i}(T_i) = \sum_{k \in \chi_{ii'}} \rho(||\mathbf{x}^k_i - T\mathbf{x}^k_{i'}||_{\Sigma}) + \lambda_h ||q(\Pi_i) - q(T_i^{-T}\Pi_{i'})||^2_2,
    \label{eq:loss_plane}
\end{equation}
where $\chi_{ii'}$ is the set of keypoint matches between planes $i$ and $i'$. $\mathbf{x}^k_i$ and $\mathbf{x}^k_{i'}$ are homogeneous coordinates $\left[x, y, z, 1\right]$ of the two matched keypoints. $\rho\left(\cdot\right)$ is the robust Huber error function \cite{mur2017orb}. $\lambda_h$ is the parameter to weight the error between the Hessian form of planes. $q(\Pi) = \left[\arctan(\frac{n_x}{n_z}), \arctan(\frac{n_y}{n_z}),d\right]$ avoids over-parametrisation of the Hessian form \cite{Li2021PlanarSLAM}.

To cluster planes with similar motions, we introduce a score $b_{ij} \in [0,1]$ for each pair of neighbouring planes $i$ and $j$ in the current frame. $b_{ij}$ represents the probability that the motion of planes $i$ and $j$ can be modelled by the same rigid transformation. 
We further introduce a new formulation based on planes to jointly estimate motion of planes and merge planes into rigid bodies: 
\begin{equation}
\begin{aligned}
    &\min_{^{ego}\mathbf{T}, \mathbf{b}} \sum_{i=1}^P e_i(T_i) + \lambda_1\sum_{(i, j) \in E_p}b_{ij}f(T_i, T_j) - \lambda_2 \sum_{(i, j) \in E_p} b_{ij}, \\
    &\text{ s.t.\ $  b_{ij} \in [0,1] \  \forall i, j$}.
    \label{eq:loss_multimotion}
\end{aligned}
\end{equation}
$^{ego}\mathbf{T}$ is the set of egocentric transformations $\{T_1, \cdots, T_P\}$ for all planes in the current frame. $\mathbf{b} = \{b_{ij}|(i, j) \in E_p\}$. $E_p$ is the connectivity graph of planes in the current frame and $(i, j) \in E_p$ means that planes $i$ and $j$ are connected in space. The first term $e_i(T_i)$ is introduced in \Cref{eq:loss_plane}. In the second term, we propose $f(T_i, T_j) = \left[e_i(T_j) + e_j(T_i) \right] - \left[e_i(T_i) + e_j(T_j)\right]$ to quantify the error between two planes with egocentric motion $T_i$ and $T_j$ respectively. The last term penalises the model complexity by maximising the sum of probabilities that neighbouring planes have similar motions.

The novelty of the formulation is that we treat each individual plane as a motion hypothesis and estimate the likelihood $\mathbf{b}$ of any two neighbouring hypotheses having the same motion. This is in contrast to MVO \cite{judd2018multimotion}, which discretely decides whether two motion hypotheses are merged or not. 

To minimise \Cref{eq:loss_multimotion}, $^{ego}\mathbf{T}$ and $\mathbf{b}$ are decoupled. We firstly initialise all egocentric motions $T_i$ to identity and all scores $b_{ij}$ to 0. Then, at each iteration, we fix $\mathbf{b}$ and find optimal $^{ego}\mathbf{T}$ by optimising each transformation independently. $\mathbf{b}$ is analytically solved subsequently by fixing the optimised transformations. After minimisation, we set a threshold $\hat{b}=0.9$ and merge planes $i$ and $j$ if $b_{ij} > \hat{b}$. We therefore acquire $M$ planar rigid bodies and use RANSAC to estimate their prior egocentric motions $\{\tilde{\xi}_1, \cdots,  \tilde{\xi}_M\}$ respectively. However, since the dynamic objects can occupy the major part of the images, we still need to decide which planar rigid body belongs to the static background.  

\subsection{Joint camera tracking and background segmentation}
We jointly track the camera motion and segment the static background based on a hierarchical representation of planes and non-planar super-pixels. This representation is more efficient in planar environments than uniformly sampled clusters used in previous work \cite{scona2018staticfusion,long2021rigidfusion}. In addition, compared to RigidFusion \cite{long2021rigidfusion}, our method only requires the camera motion prior. The dynamic planar objects are detected by their different rigid motions compared to the camera motion 
while dynamic non-planar areas are removed by their high residuals. To achieve it, we propose to minimise a combined formulation that consists of three energy terms:
\begin{equation}
    \min_{\xi_c, \bm{\gamma}} \ R(\xi_c, \bm{\gamma}) + G(\xi_c, \bm{\gamma}) + H(\xi_c) \quad  \text{ s.t.\ $ \gamma_i \in [0,1] \  \forall \, i$}, 
    \label{eq:loss_static_dynamic}
\end{equation}
where $\bm{\gamma}$ is the full set of probabilities that each plane or super-pixel is static. $\xi_c \in \mathfrak{se}(3)$ is the camera ego-motion in the world frame. Importantly, planes that belong to the same planar rigid body are assigned with independent scores $\gamma$. The first term $R(\xi_c, \bm{\gamma})$ aligns the static rigid body using weighted intensity and depth residuals. The second term $G(\xi_c, \bm{\gamma})$ segments dynamic objects by either different motions or high residuals and maintains segmentation smoothness. The last regularisation term $H(\xi_c)$ adds a soft constraint on the camera motion.

\subsubsection{Residual term}
Following the previous work \cite{scona2018staticfusion, long2021rigidfusion}, we consider image pairs ($I_{t-1}, D_{t-1}$) and ($I_t, D_t$) from two consecutive frames. For a pixel $p$ with coordinate $\mathtt{x}^p_{t} \in \mathbb{R}^2$ in the current frame $t$, the intensity residual $r_I^p(\xi)$ and depth residual $r_D^p(\xi)$ against the previous frame under motion $\xi$ are given by:
\begin{align}
    r_I^p(\xi) &= I_{t-1}\left(\mathcal{W}(\mathtt{x}^p_{t}, \xi)\right) - I_t\left(\mathtt{x}^p_{t}\right)\\
    r_D^p(\xi) &= D_{t-1}\left(\mathcal{W}(\mathtt{x}^p_{t}, \xi)\right) - |T(\xi)\pi^{-1}(\mathtt{x}^p_{t}, D_t\left(\mathtt{x}^p_{t})\right)|_z \ ,
\end{align}
where $\pi: \mathbb{R}^3 \rightarrow \mathbb{R}^2$ is the camera projection function and $|\cdot|_z$ returns the $z$-coordinate of a 3D point. The image warping function $\mathcal{W}$ is:
\begin{align}
    \mathcal{W}(\mathtt{x}^p_{t}, \xi) = \pi\left(T(\xi)\pi^{-1}(\mathtt{x}^p_t, D_t(\mathtt{x}^p_t))\right),
\end{align}
which provides the corresponding coordinate $\mathtt{x}^p_{t-1}$ in the previous frame. Similar to SF, the weighted residual term is:
\begin{align}
    R(\xi_c, \bm{\gamma}) = \sum_{p=1}^{N} \gamma_{i(p)}[F(\alpha_I w_{I}^{p}r_{I}^{p}( \xi_c)) + F(w_{D}^{p}r_{D}^{p}(\xi_c))] \ ,
    \label{eq:loss_motion}
\end{align}
where $N$ is the number of pixels with a valid depth value and $i(p) \in [1, P+S]$ indicates the index of the segment that contains the pixel $p$. $\alpha_I$ is used to weight the intensity residuals. The Cauchy robust penalty:
\begin{equation}
    F(r) = \frac{c^2}{2} \log\left(1 + \left(\frac{r}{c}^2\right)\right)
\end{equation}
is used to control robustness of minimisation and $c$ is the inflection point of $F(r)$. Compared to SF, which assigns scores to each cluster, we represent the image a combination of planes and super-pixels.

\subsubsection{Segmentation term}
The objective of $G(\xi_c, \bm{\gamma})$ is to detect dynamic planar rigid bodies by their motions and dynamic non-planar super-pixels by their high residuals. $G(\xi_c, \bm{\gamma})$ is computed by the sum of three items:
\begin{equation}
    G(\xi_c, \bm{\gamma}) = \lambda_{p}G_p(\xi_c, \bm{\gamma}) + \lambda_{np}G_{np}(\bm{\gamma}) + \lambda_{r}G_{r}(\bm{\gamma}),
\end{equation}
where $\lambda_{p}$, $\lambda_{np}$ and $\lambda_{r}$ are parameters to weight different items. The first term $G_p(\xi_c, \bm{\gamma})$ classifies planes as dynamic when their egocentric motions are different from the camera motion $\xi_c$:
\begin{equation}
    G_p(\xi_c, \bm{\gamma}) = \sum_{i=1}^{P} \gamma_i \rho(||\xi_c - \tilde{\xi}_{m(i)}||_2^2),
    \label{eq:seg_planar}
\end{equation}
where $m(i)$ is the planar rigid body that contains the plane $i$. $\tilde{\xi}_{m(i)}$ is the egocentric motion prior of the $m$-th planar rigid body and Huber cost function $\rho(\cdot)$ is used to robustly control the error. 

The second term $G_{np}(\bm{\gamma})$ handles non-planar dynamic areas. We follow StaticFusion and assume they have a significantly higher residual under the camera motion:
\begin{equation}
    G_{np}(\bm{\gamma}) = F(\hat{c})\sum_{i=P+1}^{P+S} (1 - \gamma_i) N_i,
    \label{eq:seg_non_planar}
\end{equation}
where we only consider super-pixels in non-planar area and $N_i$ is the number of pixels with valid depth in the $i$-th super-pixel. The threshold $\hat{c}$ is chosen as the average residual over all $S$ super-pixels.
 
The last term $G_r(\bm{\gamma})$ maintains the spacial smoothness of segmentation $\bm{\gamma}$ for both planar and non-planar areas by encouraging neighbour areas to have close scores:
\begin{equation}
    G_{r}(\bm{\gamma}) = \sum_{(i, j) \in E_{p}}b_{ij}(\gamma_i - \gamma_j)^2 + \sum_{(i, j) \in E_{np}} (\gamma_i - \gamma_j)^2,
    \label{eq:seg_connectivity}
\end{equation}
where $E_p$ and $E_{np}$ is the connectivity graph for planes and non-planar super-pixels respectively. $b_{ij}$ is directly acquired from the minimisation of \Cref{eq:loss_multimotion}. This means that rather than directly assigning the same score $\gamma$ to planes that belong to the same rigid body, we encourage them to have a close score $\gamma$.

\subsubsection{Motion regularisation term}
We add a soft constraint on the camera motion $\xi_c$ based on the motion prior $\tilde{\xi}_c$:
\begin{equation}
    H( \xi_c) = \lambda_c (1-\alpha_s) \rho (||\xi_c - \tilde{\xi}_c||_2^2),
    \label{eq:loss_priorpose}
\end{equation}
where $\alpha_s \in [0, 1]$ is the proportion between the the number of pixels that are associated to the static background over the total number of pixels with valid depth reading. This means that we rely more on the camera motion prior when the dynamic objects occupy a higher proportion of the image view. The robust Huber cost function $\rho(\cdot)$ is used to handle large potential drifts in the camera motion prior.

The solver of \Cref{eq:loss_static_dynamic} is based on StaticFusion and a similar coarse-to-fine scheme is applied to directly align dense images. Specifically, we create an image pyramid for each incoming RGB-D image and start the optimisation from the coarsest level. The results acquired in the intermediate level are used to initialise the next level, to allow correct convergence. We also decouple the camera motion $\xi_c$ and $\bm{\gamma}$ for more efficient computation. Concretely, we initialise the camera motion $\xi_c$ as identity and all $\bm{\gamma}$ to 1. For each iteration, we first fix $\bm{\gamma}$ and find the optimal $\xi_c$. The closed-form solution for $\bm{\gamma}$ is then obtained by fixing $\xi_c$. The solution for the previous iteration is used to initialise the current iteration.

\subsection{Background reconstruction and camera pose refinement}
In the current frame $t$, after the minimisation of \Cref{eq:loss_static_dynamic}, we acquire the optimised camera motion $\hat{\xi}_c$ and the static parts of intensity and depth images $(I^s_t, D^s_t)$. These images are used to reconstruct the static background and refine the camera pose $\xi_c$ using frame-to-model alignment. Concretely, we render an image pair $(I^r_{t-1}, D^r_{t-1})$ from the current static background model at the previous camera pose. The rendered image pair $(I^r_{t-1}, D^r_{t-1})$ is directly aligned with $(I^s_t, D^s_t)$ by minimising
\begin{equation}
    \min_{\xi_c} \ R(\xi_c, \bm{\gamma} = \mathbf{1}) + H(\xi_c).
    \label{eq:loss_frame_model_alignment}
\end{equation}
The first term $R(\xi_c, \bm{\gamma} = \mathbf{1})$ is the same as \Cref{eq:loss_motion} but the $\bm{\gamma}$ is fixed to $\mathbf{1}$ because the input should only contain the static background. We append $H(\xi_c)$ in \Cref{eq:loss_priorpose} as a soft-constraint for the frame-to-model alignment and $\alpha_s$ is estimated from pixel-wise dynamic segmentation $\Gamma_t$. Since we have already solved $\Cref{eq:loss_static_dynamic}$, we directly start from the finest level of the image pyramid and initialise the solver with the camera pose $\hat{\xi}_c$ for the solver of \Cref{eq:loss_frame_model_alignment}. The refined camera pose is used to fuse the static images $(I^s_t, D^s_t)$ with the surfel-based 3D model as described in SF \cite{scona2018staticfusion}.

\subsection{Planar objects tracking}
After removing the static planes, we further track dynamic planar rigid bodies independently. This is different to our previous work RigidFusion \cite{long2021rigidfusion} which models the whole dynamic component with a single rigid transformation. For each dynamic planar rigid body $m$, we match it to the previous dynamic rigid bodies using the plane association and estimate the egocentric motion. If all the currently associated planes are static in the previous frame, we detect the dynamic planar rigid body $m$ as a new object and the initial pose of the object relative to the camera frame is denoted as $T_{init}$. If the initial time of frame for an object is $t_0$, the object pose in the object's initial frame can be acquired by \cite{wang2020dymslam}:
\begin{align}
    T_{O^m_{t_0} O^m_t} = T_{C_{t_0}C_{t}} {^m}T_{C_{t_0}C_t}T_{init}^{-1}.
\end{align}

\section{Evaluation}
\subsection{Setup}
\begin{figure}[tb]
    \centering
    \setlength{\belowcaptionskip}{0cm}
    \subfloat[Mobile manipulator Ada]{\includegraphics[height=4cm]{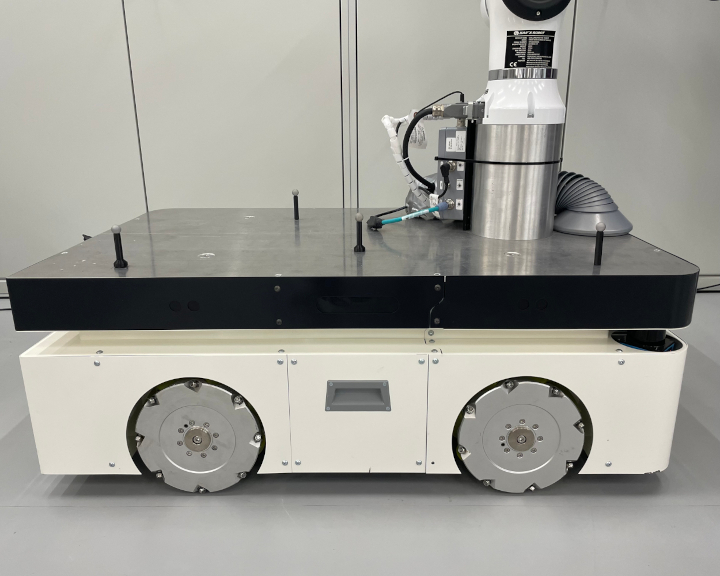} \label{fig:ada}}
    \subfloat[Object  1]{\includegraphics[height=4cm]{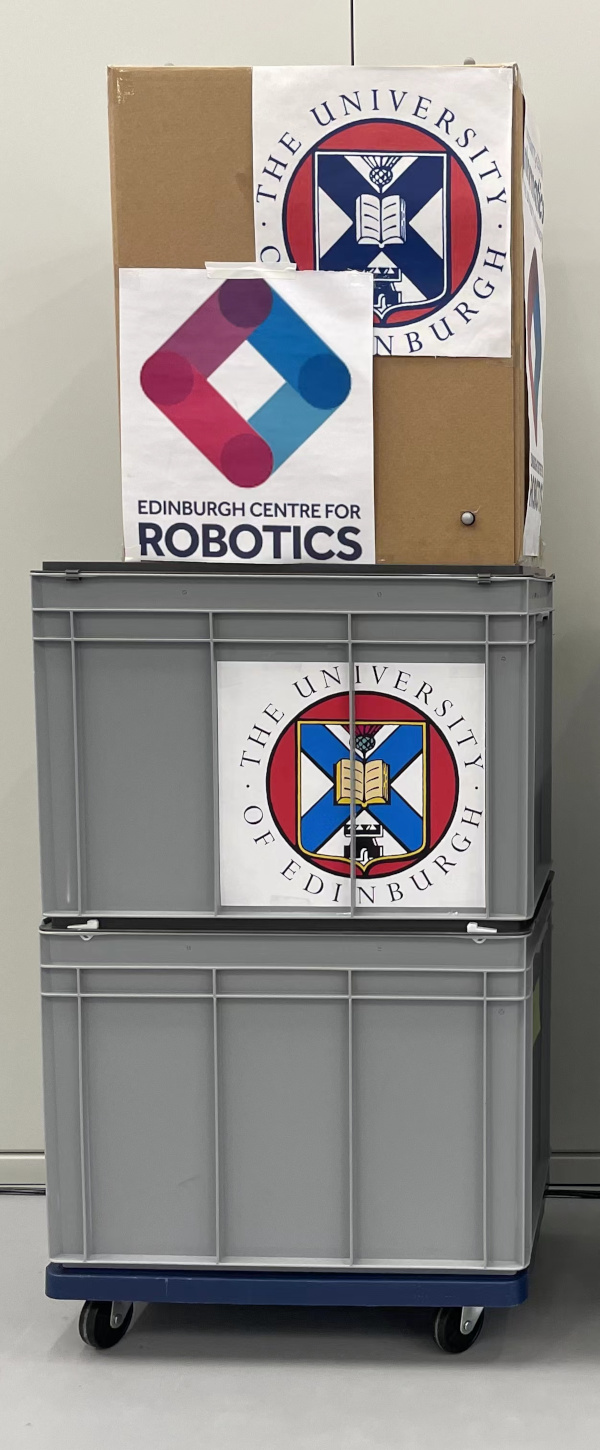} \label{fig:object1}}
    \subfloat[Object 2]{\includegraphics[height=4cm]{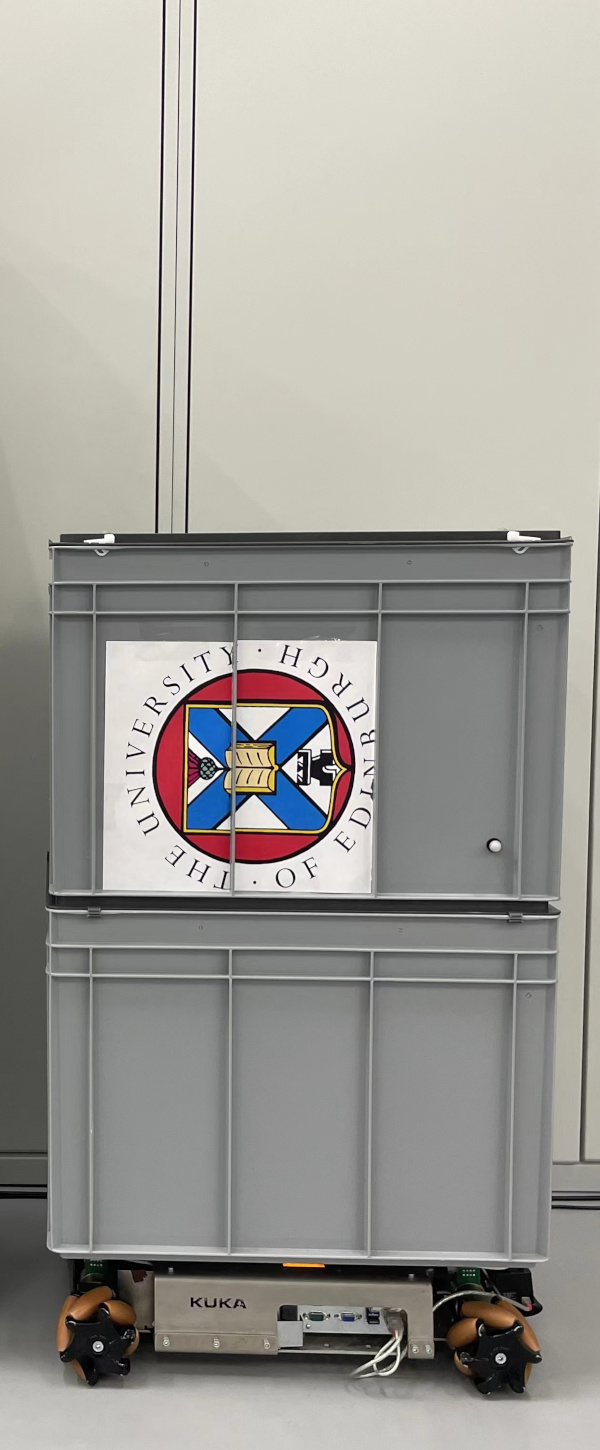} \label{fig:object2}}
    \caption{(a) An omnidirectional wheeled platform with Vicon markers. (b) The first rigid object is put on a board with wheels and moved by a human. (c) The second rigid object is put on the youBot and is controlled remotely.}
    \label{fig:setup}
\end{figure}

The sequences for evaluation are collected with an Azure Kinect DK RGB-D camera which is mounted on an omnidirectional robot (\Cref{fig:ada}). The camera produces RGB-D image pairs with a resolution of 1280 x 720 at 30 Hz. The images are down-scaled and cropped to 640 x 480 (VGA) to accelerate the speed of pre-processing (\Cref{fig:pipeline}), such as super-pixel and plane extraction. In the solver of \Cref{eq:loss_static_dynamic} and (\ref{eq:loss_frame_model_alignment}), we further down-scale images to 320 x 240 (QVGA).

The dynamic objects are created from stacked boxes and are either moved by humans or via a remotely controlled KUKA youBot (\Cref{fig:setup}). The ground truth trajectories of the camera and objects are collected using a Vicon system by attaching Vicon markers on the camera and dynamic objects. The camera motion prior is acquired by adding synthetic drift on camera ground truth trajectories with a magnitude of around 7 cm/s (trans.) and 0.4 rad/s (rot.).

For quantitative evaluation, we estimate the absolute trajectory error (ATE) and the relative pose error (RPE) \cite{sturm2012benchmark} against the ground truth. The proposed method is compared with PlanarSLAM (PS) \cite{Li2021PlanarSLAM}, EM-Fusion (EMF) \cite{Strecke2019EMFusion}, Joint-VO-SF (JF) \cite{jaimez2017fast}, StaticFusion (SF) \cite{scona2018staticfusion}, Co-Fusion (CF) \cite{runz2017co} and RigidFusion (RF) \cite{long2021rigidfusion}. We additionally provide the camera motion prior with drift to CF as the variant CF$^{\mathbf{*}}$. The original RF uses motion priors for both camera and object. Here we only provide RF with the camera motion prior and denote it as RF$^{*}$, while our method with the camera motion prior is denoted as ours$^{*}$.

\begin{figure}[t]
    \setlength{\belowcaptionskip}{0cm}
    \centering
    \includegraphics[width=\linewidth]{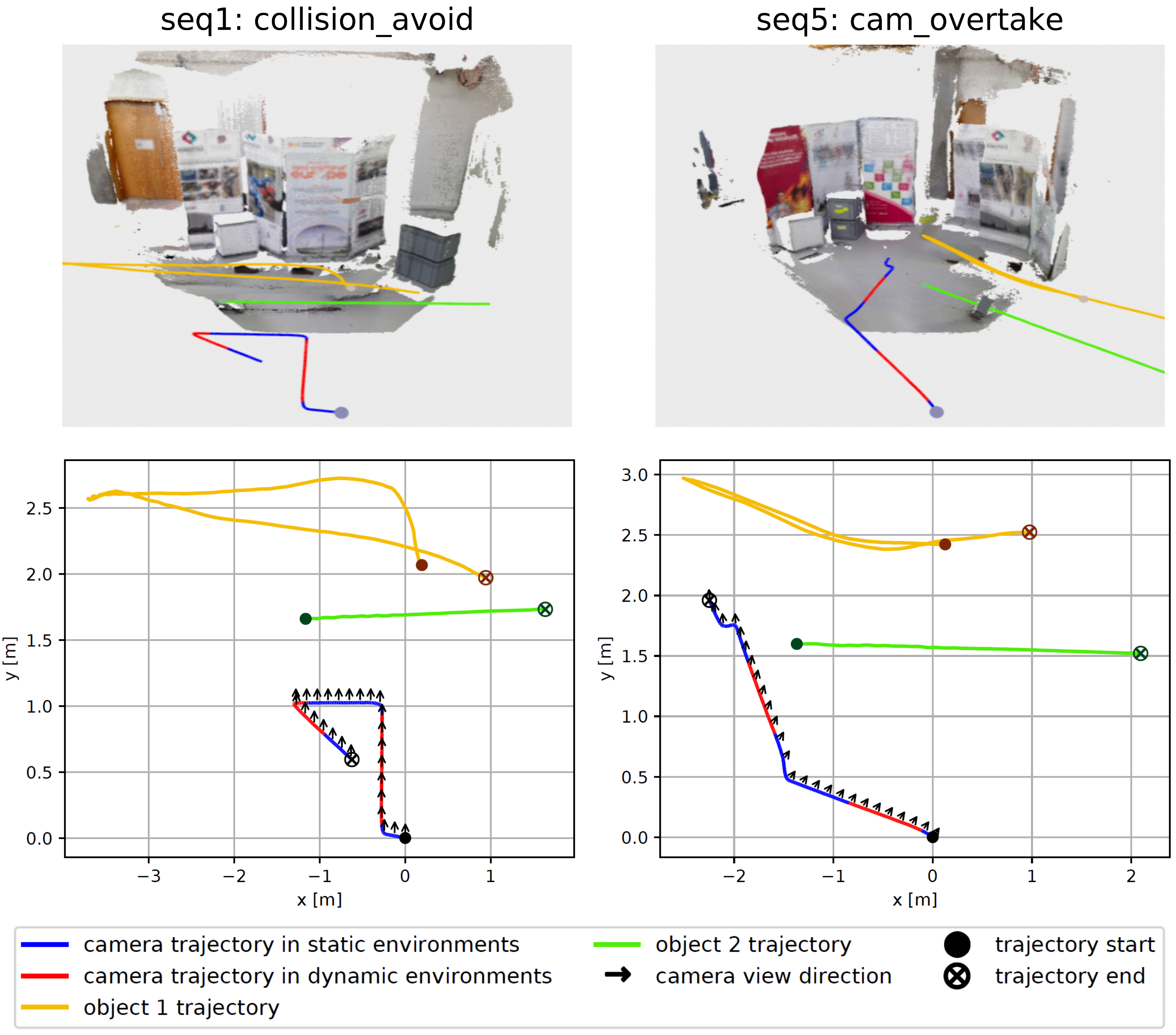}
    \caption{The ground truth trajectories of camera and dynamic objects in both 2D and 3D perspectives. Trajectories of humans are not plotted. The red segment of camera trajectories represents the part when there are moving objects in the camera view, while the blue segment means the camera moves in static environments. We mark trajectories' start position with a black
    solid dot and end position with a circle-cross marker. The black arrows point to the direction of camera view.}
    \label{fig:motion}
\end{figure}

We collect eight sequences with various camera and object movements in different planar environments. For example, in the \textit{seq1}, a human moves the taller box to clear way for both the robot and the other object so that the potential collision can be avoided, while in the \textit{seq5}, the robot tries to overtake two dynamic objects ahead (\Cref{fig:motion}). All trajectories are designed such that the two dynamic objects and a human can be visible in the image at the same time and frequently occupy the major proportion of the camera view. We also run experiments on sequences \textit{sitting\_xyz} and \textit{walking\_xyz} from TUM RGB-D dataset \cite{sturm2012benchmark} which includes a large proportion of non-planar areas and denote them as \textit{seq9} and \textit{seq10} respectively.

\subsection{Camera localisation}
\begin{figure*}[t]
    \setlength{\belowcaptionskip}{-0.2cm}
    \centering
    \includegraphics[width=\linewidth]{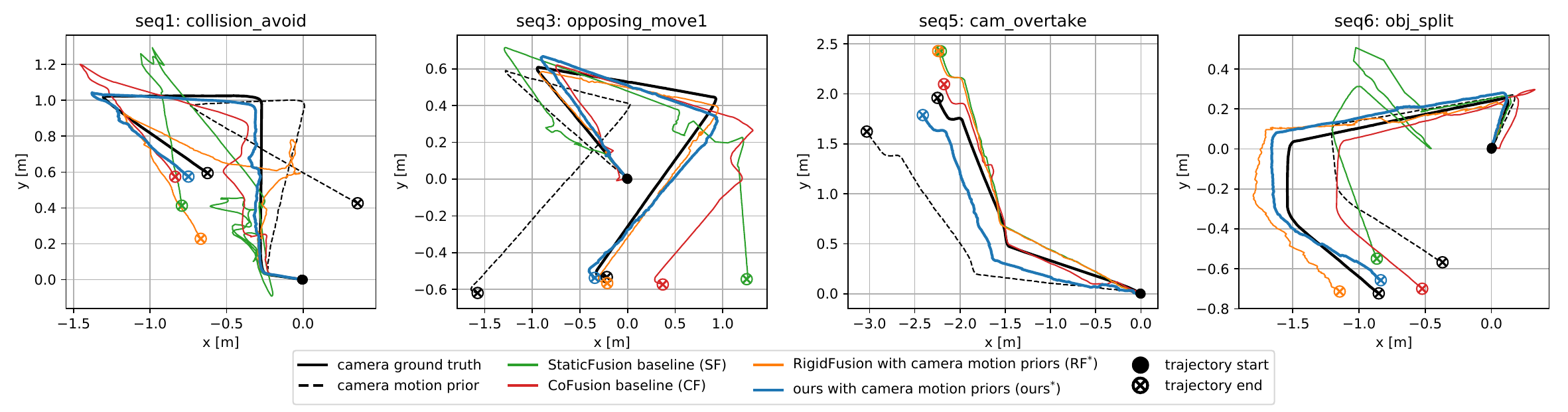}
    \caption{Visualisation of the estimated camera trajectories, camera motion prior and ground truth. The start position of all trajectories is aligned to the same position and is marked with a black solid dot. Our method (blue) achieves the lowest error compared to the ground truth (black solid) and can correct the drift of the camera motion prior (black dashed). }
    \label{fig:camera_trajectories}
\end{figure*}

We estimate the ATE root-mean-square error (RMSE) and RPE RMSE between the estimated camera trajectories and ground truth (\Cref{tab:camera_trajectories}). In planar dynamic environments (seq. 1-8), the evaluation demonstrates that our method outperforms all other state-of-the-art methods (\Cref{fig:camera_trajectories}). With the help of the camera motion prior, our method achieves the best performance and corrects the large drift of the camera motion prior. Even without the camera motion prior, we still achieve better performance than the baseline of JF, SF and CF. PS is unable to estimate the correct camera pose because there are dynamic planes in the environment while PS assumes all planes are static. Our method also outperforms EMF because the semantic segmentation method \cite{he2017mask} can only detect and segment certain categories of dynamic objects, like humans.

In non-planar dynamic environments (seq. 9-10), EMF outperforms all other methods because the dynamic humans can be directly segmented by Mask R-CNN \cite{he2017mask}. However, even without relying on semantic segmentation, our method has close performance compared to StaticFusion. This is because our method can still detect dynamic super-pixels by their high residuals under the camera motion.

\begin{table}[tb]
\centering
\begin{subtable}{\linewidth}
    \centering
\resizebox{\columnwidth}{!}{%
 \begin{tabular}{|c|c|ccccccccc|}
    \hline
    \multicolumn{1}{|c|}{} & MP & \multicolumn{9}{c|}{SLAM Method} \\
\cline{3-11}    \multicolumn{1}{|c|}{} &  &\multicolumn{1}{c|}{PS} & \multicolumn{1}{c|}{EMF} & \multicolumn{1}{c|}{JF} & \multicolumn{1}{c|}{SF} & \multicolumn{1}{c|}{CF} & \multicolumn{1}{c|}{CF$^{*}$} & \multicolumn{1}{c|}{RF$^{*}$} & \multicolumn{1}{c|}{ours} &ours$^{*}$ \\
\hline 1 & 26.7  & 38.5 & 50.6 & 30.5   & 22.9 & \textit{10.4}  & 10.2 & 16.5  &  20.1   & \textbf{4.23} \\
\hline 2 & 49.5  & 88.7 & 63.6 & 28.2    & 27.4 & 26.0 & 7.30 & 14.3 &  \textit{6.81}   & \textbf{6.32} \\
\hline 3 & 41.7  & 53.1 & 37.0  & 24.3    & 74.0 & 21.6      & 10.6  & 4.38  &  \textit{4.01}   & \textbf{3.42} \\
\hline 4 & 36.0 & 36.8 & 34.0 & 28.9   & 87.2 & \textit{18.9} & 20.3 & 8.39 &  22.6   & \textbf{8.37}  \\
\hline 5 & 16.2 & 31.4 &  14.7 & 10.3     & 13.6 & \textit{\textbf{4.73}}   &  8.35  & 14.1  &   25.2   & 6.74  \\
\hline 6 & 11.7  & 39.6 & 35.5 & 52.8     & 23.5  & 10.1 &\textbf{3.67}  & 7.57  &  \textit{8.37}   & 4.67 \\
\hline 7  & 25.5 & 19.1 &25.5  & 34.7 & 57.6      & 14.7     & 8.71 &41.3    &   \textit{6.43}  & \textbf{7.60} \\
\hline 8  & 28.4 & 46.8 & 25.6 & 26.5 &62.1 &69.8      & 18.9     & 14.2 &  \textit{\textbf{8.33}}   &10.3 \\
\hline 
\hline 9  & 273 & $\mathbf{2.15}$ & $3.7^{\dagger}$ & 11.1$^{\dagger}$ & 4.0$^{\dagger}$ & 2.7$^{\dagger}$ & 5.63 & 9.73 & 3.81 & 5.54\\
\hline 10 & 197 & 29.8 & $\mathbf{6.6}^{\dagger}$ & 87.4$^{\dagger}$ & 12.7$^{\dagger}$ & 69.6$^{\dagger}$ & 48.7 & 19.5 & 14.9 & 11.6 \\
    \hline
    \end{tabular}%
}
    \caption{Trans. ATE RMSE (cm)}
  \label{tab:camera_ate}%
\end{subtable}

\vspace{0.1cm}

\begin{subtable}{\linewidth}
    \centering
    \resizebox{\columnwidth}{!}{%
    \begin{tabular}{|c|c|ccccccccc|}
    \hline
         & MP      & \multicolumn{9}{c|}{SLAM Method} \\
\cline{3-11}    &  & \multicolumn{1}{c|}{PS} & \multicolumn{1}{c|}{EMF} & \multicolumn{1}{c|}{JF} & \multicolumn{1}{c|}{SF} & \multicolumn{1}{c|}{CF} & \multicolumn{1}{c|}{CF$^{*}$} & \multicolumn{1}{c|}{RF$^{*}$} & \multicolumn{1}{c|}{ours} & ours$^{*}$ \\\hline 
1 & 7.64& 23.8 & 43.6 & 28.4& 17.2 & \textit{7.58}  &   9.07   & 9.41 &  10.9   & \textbf{4.50} \\
\hline 2 & 7.31& 51.6 & 22.8 & 26.9& 11.2  & 12.6    & 5.61 & 3.99 &  \textit{3.58}    & \textbf{3.06} \\
\hline 3 & 7.87& 25.1 & 14.8 & 23.5     & 26.1     & 6.8    &4.22 & 7.13 &  \textit{3.11}  & \textbf{2.78}  \\
\hline 4 & 7.38& 29.9 & 26.5 & 28.2      & 64.3     & 15.9 & 15.7  & 6.13 &  \textit{14.2}  & \textbf{6.52}  \\
\hline 5 & 7.61& 25.8 & 6.31& 31.4      & \textit{\textbf{3.31}}     & 3.62& 6.34 & 3.90  &  13.5  & 4.77    \\
\hline 6 & 7.51& 17.1 &  30.6  & 25.4      & 18.1     & 7.02     &  4.67  & 4.26 &  \textit{4.38}   & \textbf{3.18}  \\
\hline 7  & 7.52 & 12.8 & 15.4 & 31.3 & 62.4      & 7.54    & 6.43  & 25.1  &   \textit{4.73}   & \textbf{4.09} \\
\hline 8  & 7.29 & 20.0 & 15.6 & 24.1 & 36.4      & 28.3     & 11.2 & 7.54  &  \textit{5.91}    & \textbf{4.41} \\
\hline 
\hline 9  & 7.36 & 3.12 & $\mathbf{2.6}^{\dagger}$ & 5.7$^{\dagger}$ & 2.8$^{\dagger}$ & 2.7$^{\dagger}$ & 3.01 & 3.48 & 2.95 & 2.98 \\
\hline 10 & 7.34 & 49.0 & $\mathbf{6.0}^{\dagger}$ & 27.7$^{\dagger}$ & 12.1$^{\dagger}$ & 32.9$^{\dagger}$ & 41.9 & 13.4 & 9.59 & 8.67 \\
    \hline
   \end{tabular}%
   }
    \caption{Trans. RPE RMSE (cm/s)}
    \label{tab:camera_rpe}%
\end{subtable}
\setlength{\belowcaptionskip}{-0.2cm}
\caption{ATE and RPE RMSE for all ten RGB-D sequences. The asterisk ($*$) symbol represents that the method uses the camera motion prior with drift and the dagger ($\dagger$) symbol means the result is taken from the original paper \cite{Strecke2019EMFusion}. Our method achieves the best performance in custom robotic sequences collected from planar environments (seq. 1-8) and estimates correct camera trajectories in TUM RGB-D dataset \cite{sturm2012benchmark} which contains a large proportion of non-planar areas (seq. 9-10).}
\label{tab:camera_trajectories}
\end{table}

\subsection{Multimotion segmentation}
\begin{figure*}[tb]
\setlength{\belowcaptionskip}{-0.5cm}
\centering
\setlength{\tabcolsep}{0pt}
\newcommand{\h}{1.76cm}
\begin{tabular}{r|cccccc|}
\textbf{} & \multicolumn{6}{c|}{\textbf{segmentation}} \\
  & \multicolumn{3}{c|}{seq4: opposing\_move2}    &\multicolumn{3}{c|}{seq8: obj\_transfer}   \\

\begin{tabular}[r]{@{}r@{}}RGB input with\\planes and\\super-pixels\\ segmentation\end{tabular}\hspace{0.2cm} &  \includegraphics[height=\h,valign=m]{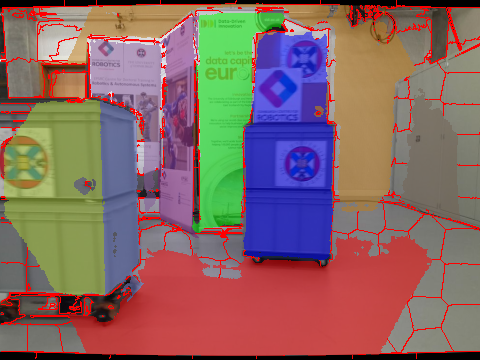}   \     & 
\includegraphics[height=\h,valign=m]{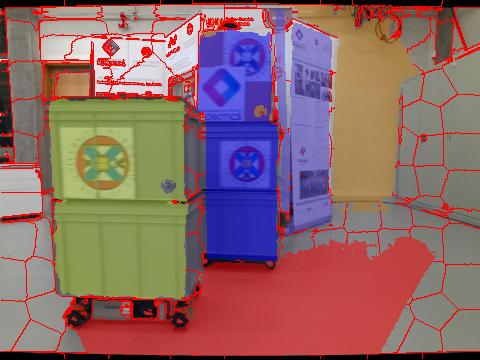}   \      & 
\includegraphics[height=\h,valign=m]{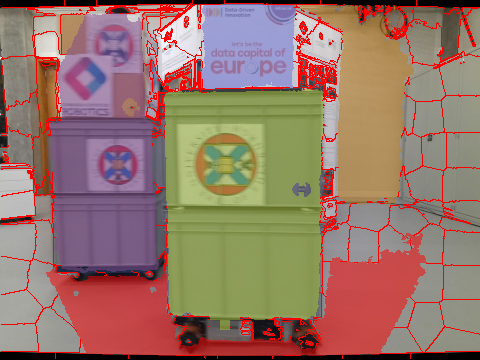}   \     &
\includegraphics[height=\h,valign=m]{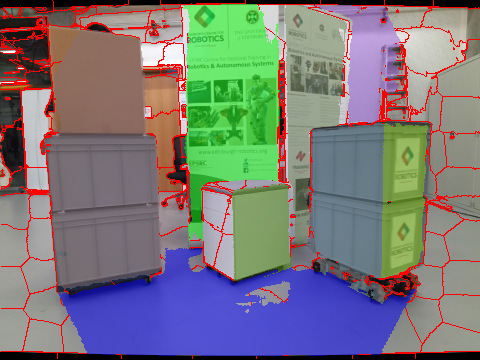}  \    & 
\includegraphics[height=\h,valign=m]{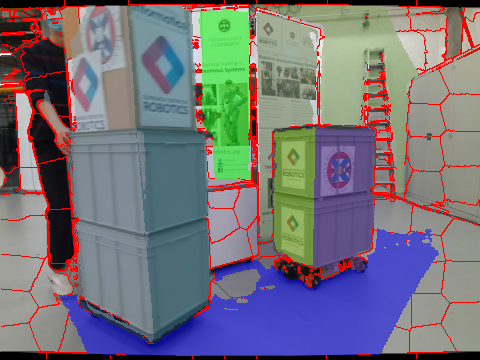}   \     & 
\includegraphics[height=\h,valign=m]{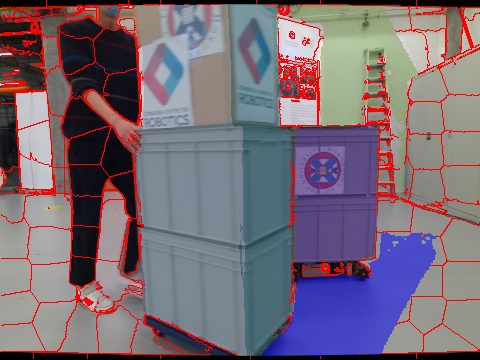} \\

\begin{tabular}[r]{@{}r@{}}SF \end{tabular}\hspace{0.2cm} &  \includegraphics[height=\h,valign=m]{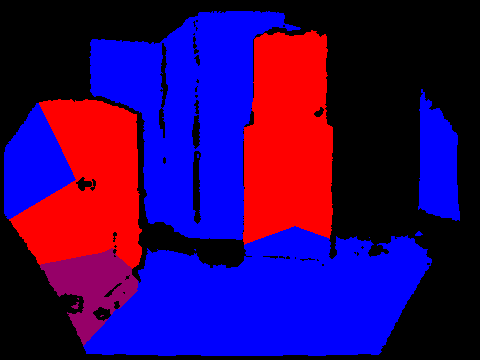}   \     & \includegraphics[height=\h,valign=m]{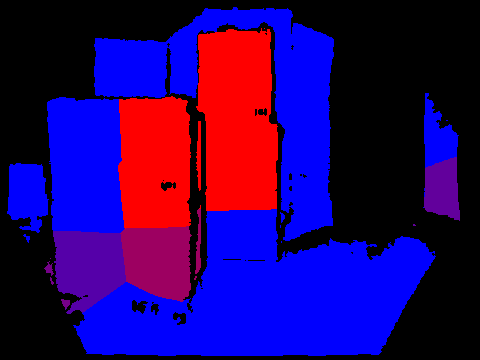}   \      & \includegraphics[height=\h,valign=m]{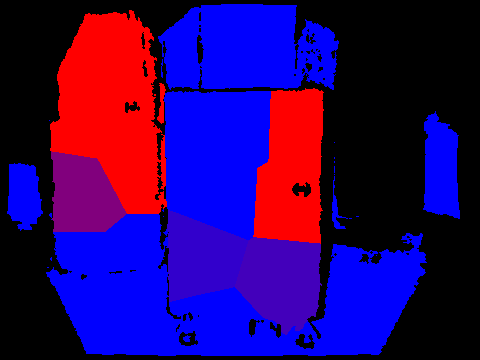}    \    &
\includegraphics[height=\h,valign=m]{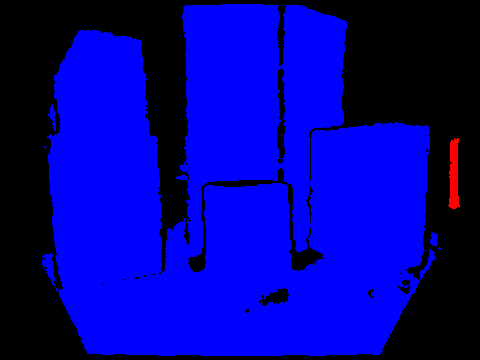}   \     & \includegraphics[height=\h,valign=m]{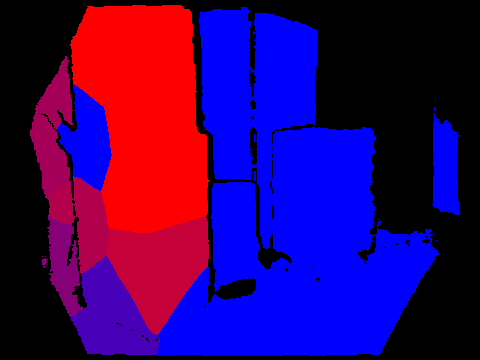}    \    & \includegraphics[height=\h,valign=m]{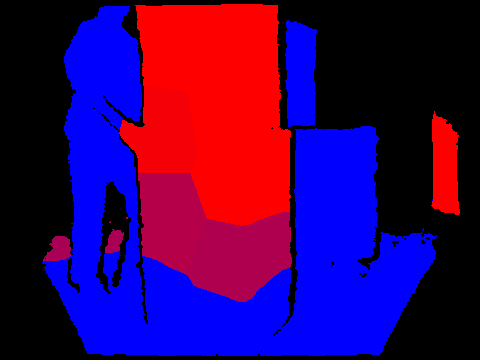}  \\

\begin{tabular}[r]{@{}r@{}}RF$^*$ \end{tabular}\hspace{0.2cm}     & \includegraphics[height=\h,valign=m]{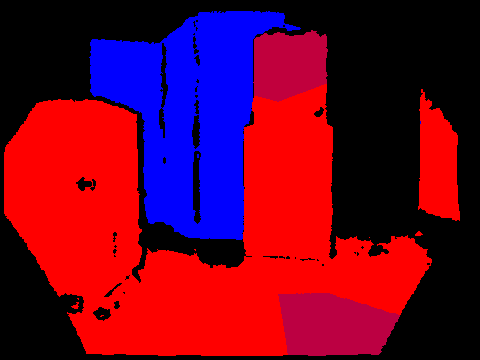}   \     & \includegraphics[height=\h,valign=m]{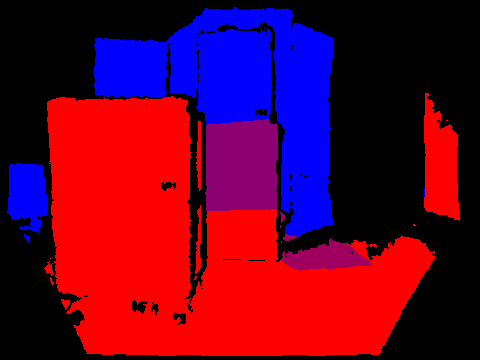}   \      & \includegraphics[height=\h,valign=m]{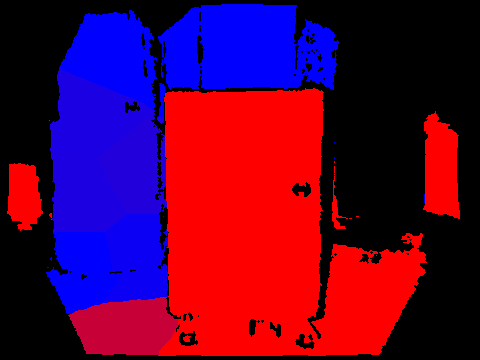}    \    &
\includegraphics[height=\h,valign=m]{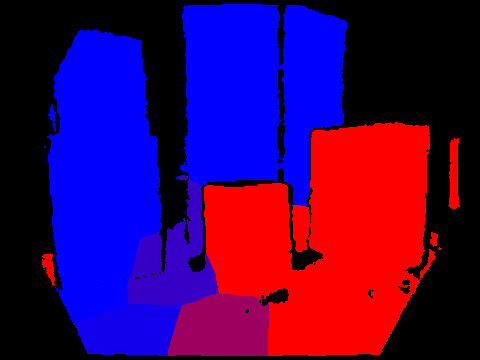}   \     & \includegraphics[height=\h,valign=m]{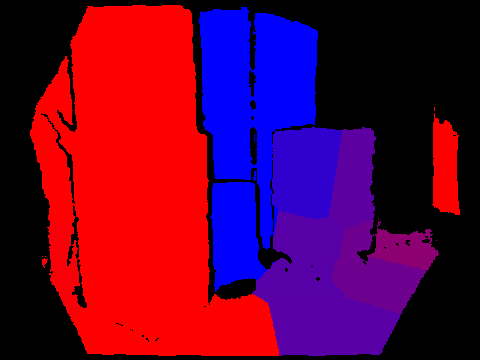}    \    & \includegraphics[height=\h,valign=m]{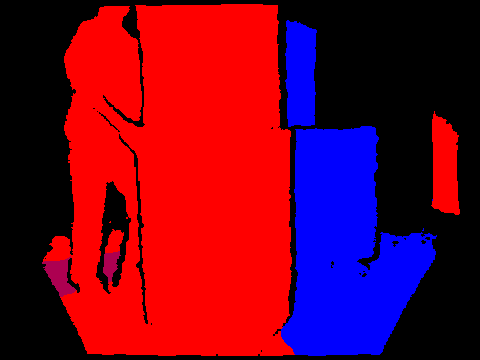}  \\

CF\hspace{0.2cm} &
\includegraphics[height=\h,valign=m]{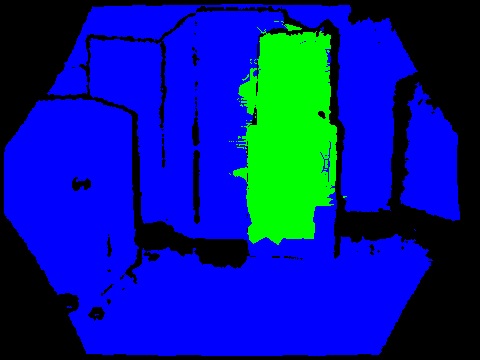}   \     & \includegraphics[height=\h,valign=m]{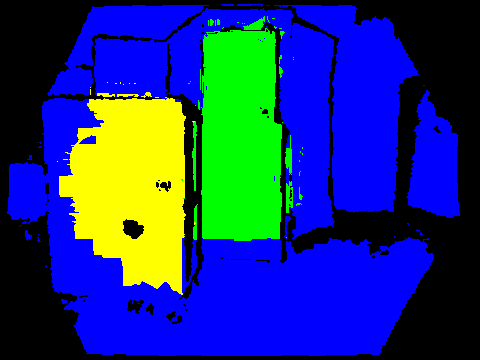}   \      & \includegraphics[height=\h,valign=m]{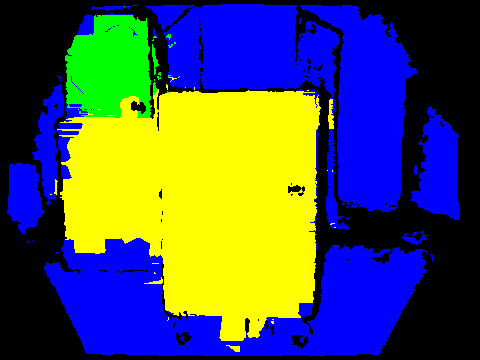}    \    &
\includegraphics[height=\h,valign=m]{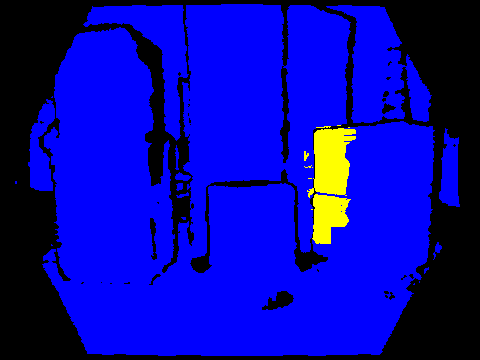}   \     & \includegraphics[height=\h,valign=m]{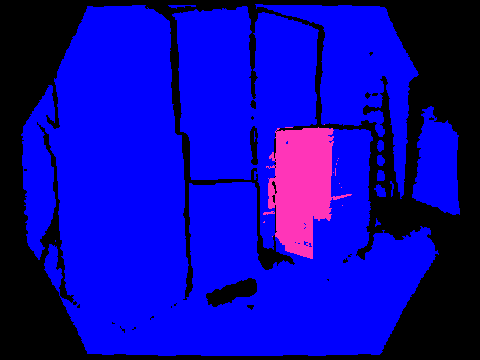}    \    & \includegraphics[height=\h,valign=m]{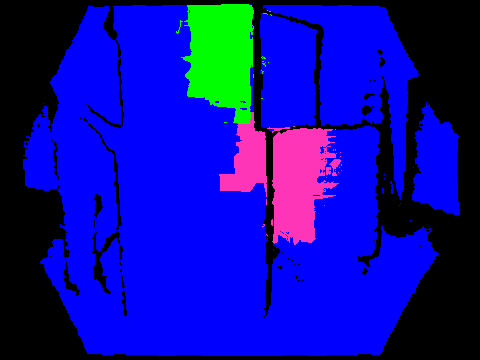}  \\

ours$^*$\hspace{0.2cm}    & \includegraphics[height=\h,valign=m]{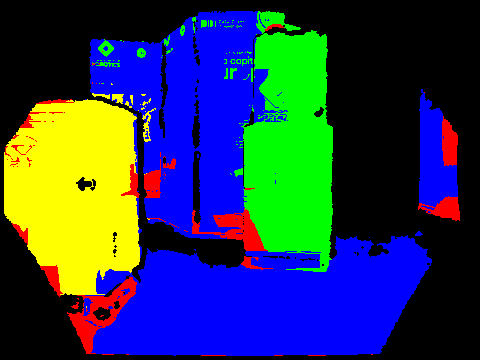}   \     & \includegraphics[height=\h,valign=m]{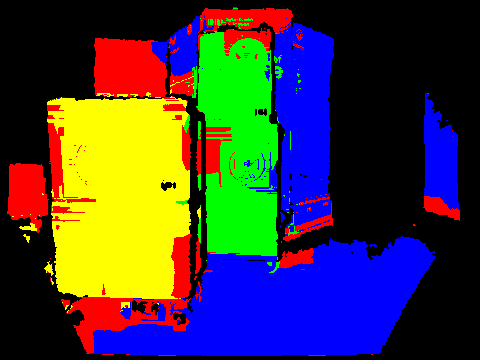}   \      & \includegraphics[height=\h,valign=m]{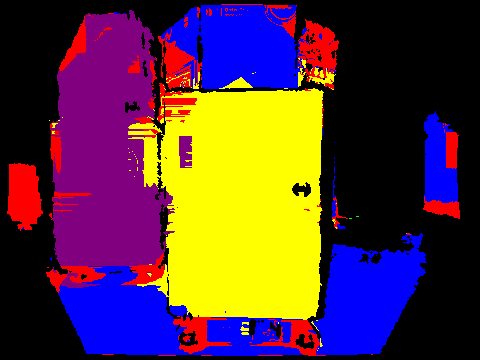}    \    &
\includegraphics[height=\h,valign=m]{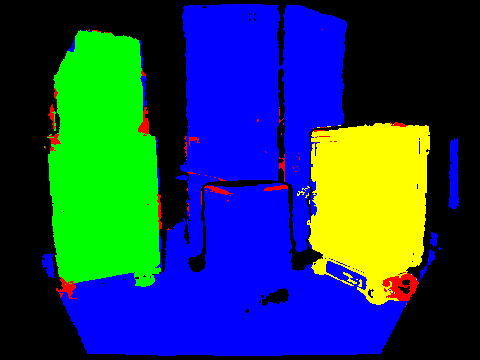}   \     & \includegraphics[height=\h,valign=m]{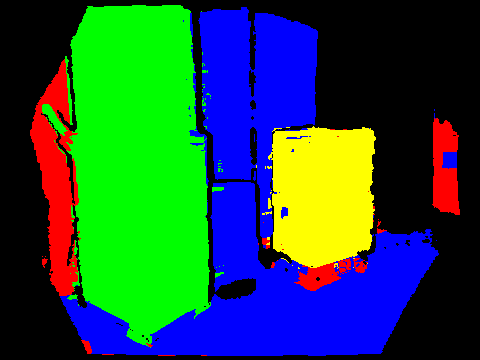}    \    & \includegraphics[height=\h,valign=m]{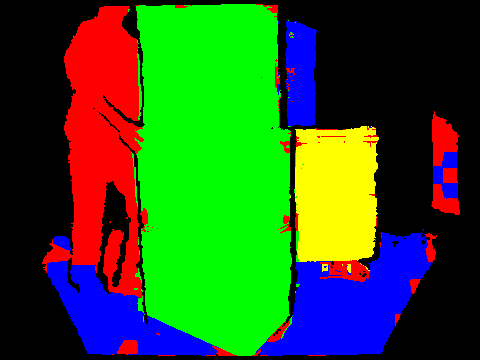}  \\  
\end{tabular}
\caption{Segmentation result of the static background and dynamic objects. We visualise the input RGB images with the segmentation of planes and super-pixels in the first row. In all four methods, the static part is marked by blue. In SF and RF, we use red to represent dynamic parts. In CF, we use different colours to show different objects. In our method, the non-planar dynamic areas are marked by red, the planar rigid objects are marked by other colours. Results show that only our method can segment multiple dynamic objects correctly and is robust to large occlusion. }
\label{fig:segmentation}
\end{figure*}

\begin{figure}[tbh]
    \centering
    \setlength{\tabcolsep}{0pt}
    \begin{tabular}{ccccc}
    \rotatebox[origin=c]{90}{RGB}\hspace{0.1cm}     &
    \includegraphics[width=0.23\linewidth,frame,valign=m]{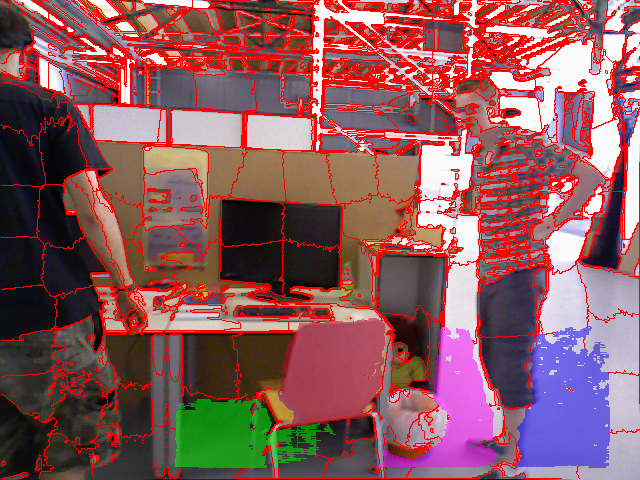} &
    \includegraphics[width=0.23\linewidth,frame,valign=m]{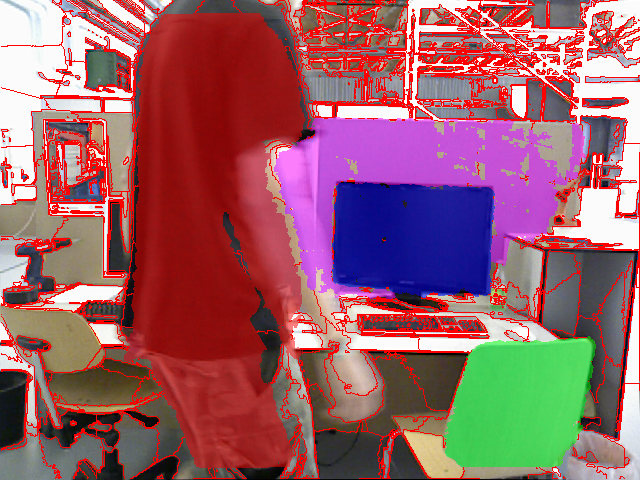}&
    \includegraphics[width=0.23\linewidth,frame,valign=m]{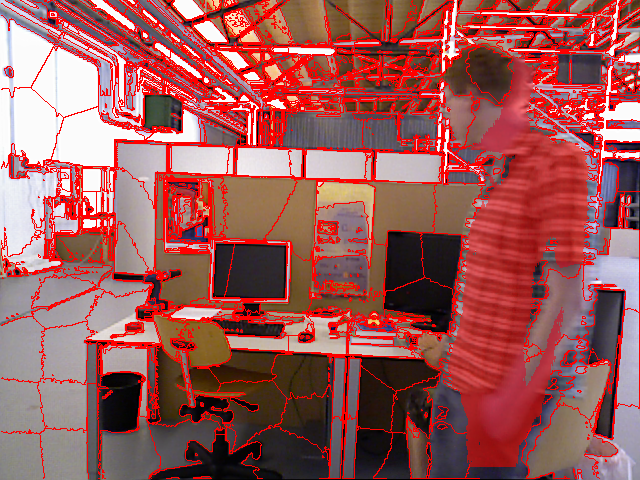}&
    \includegraphics[width=0.23\linewidth,frame,valign=m]{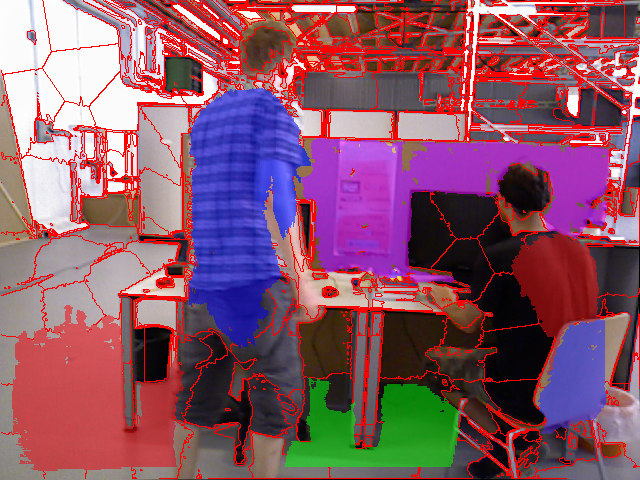}        
    \\
    \rotatebox[origin=c]{90}{SF}\hspace{0.1cm}    & \includegraphics[width=0.23\linewidth,frame,valign=m]{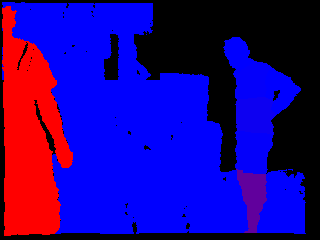} &
    \includegraphics[width=0.23\linewidth,frame,valign=m]{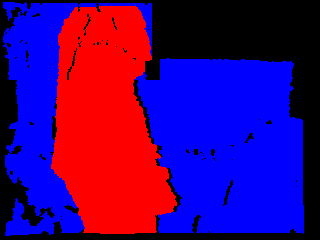}&
    \includegraphics[width=0.23\linewidth,frame,valign=m]{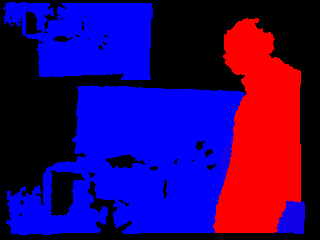}&
    \includegraphics[width=0.23\linewidth,frame,valign=m]{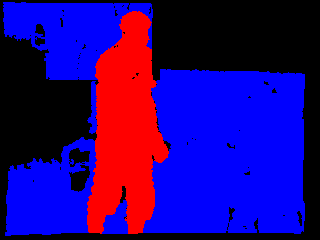}   
    \\
    \rotatebox[origin=c]{90}{ours}\hspace{0.1cm}     & \includegraphics[width=0.23\linewidth,frame,valign=m]{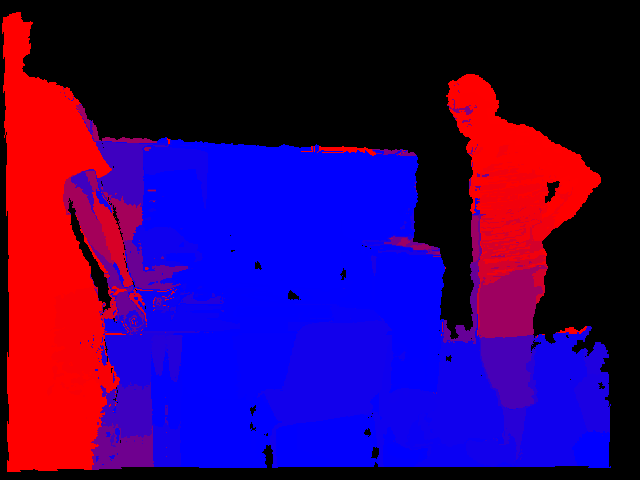} &
    \includegraphics[width=0.23\linewidth,frame,valign=m]{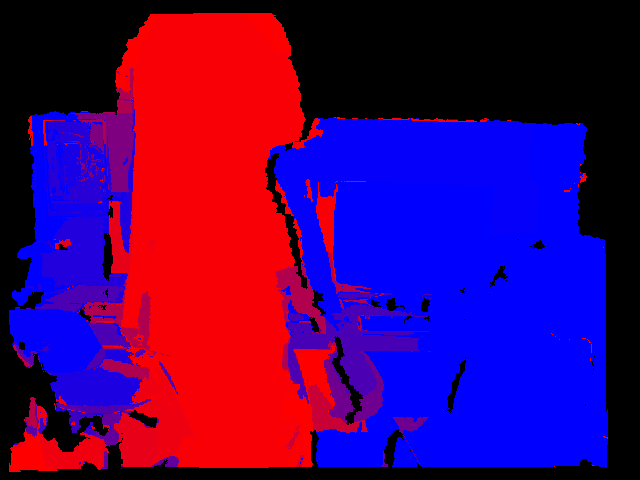}&
    \includegraphics[width=0.23\linewidth,frame,valign=m]{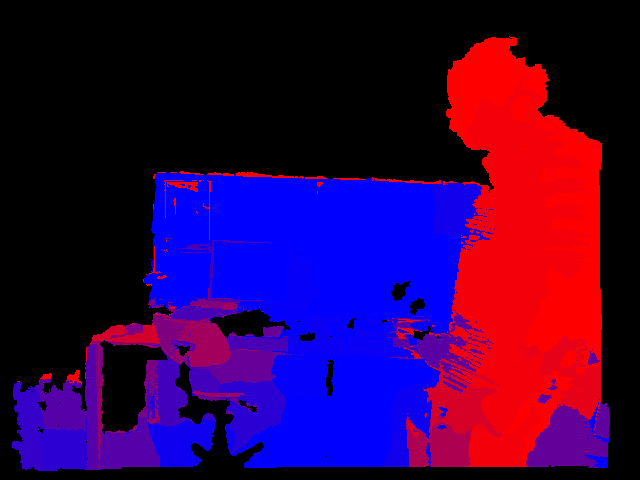}&
    \includegraphics[width=0.23\linewidth,frame,valign=m]{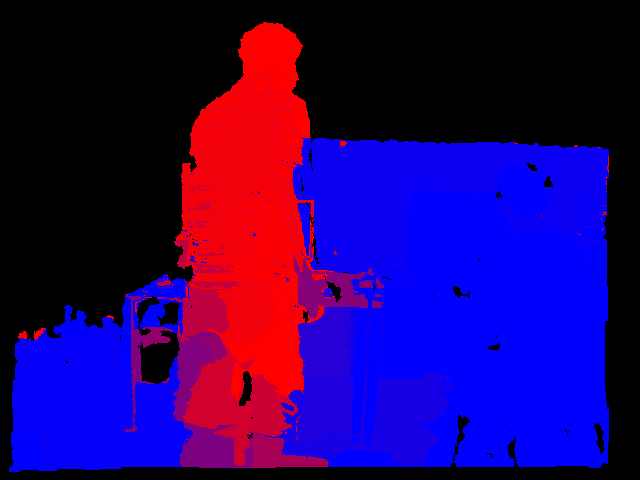}  
    \end{tabular}
    \caption{Static/dynamic segmentation results on the \textit{walking\_xyz} sequence \cite{sturm2012benchmark}. The first row shows the RGB images with segmentation of planes and super-pixels. Our method achieves close segmentation performance to SF in non-planar environments.}
    \label{fig:tum}
\end{figure}

For planar environments, we visualise the segmentation results of our method and compare them with SF, RF$^*$ and CF (\Cref{fig:segmentation}). SF is unable to detect all dynamic objects because they as a whole occlude a large proportion of the camera view, while RF$^*$ tends to classify parts of the static background as dynamic. Both CF and our method can further distinguish between different dynamic objects. However, the segmentation of CF is not complete and CF tends to have a delay when detecting a new object. We use two different colours (green and purple) to represent that our method treats the taller object as a new one after it passes behind the front object. In non-planar environments, our method can still provide correct binary segmentation of the static and dynamic objects (\Cref{fig:tum}). However, we are unable to segment and track different non-planar dynamic objects independently.

\subsection{Background reconstruction}
\begin{figure}[tbh]
    \centering
    \begin{tabular}{l|l}
    \begin{tikzpicture}
    \node[anchor=south west,inner sep=0]{\includegraphics[height=3cm]{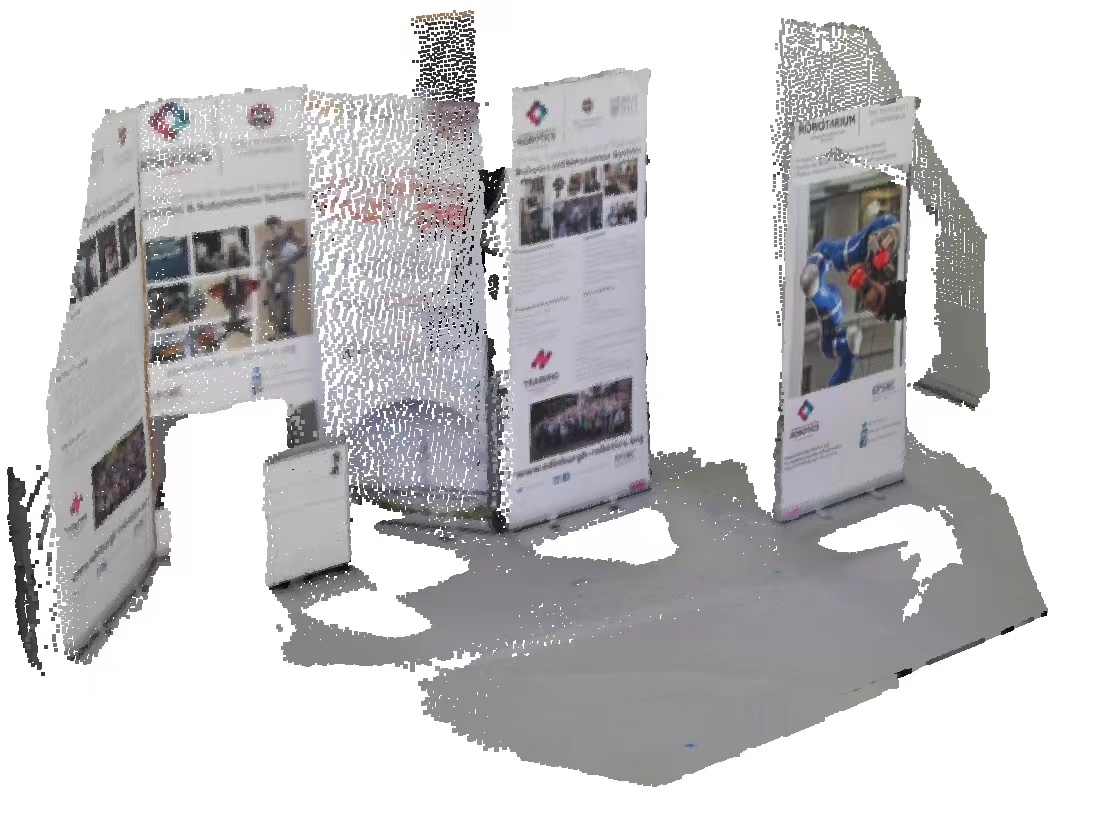}};
    \node[anchor=south west]{\textbf{Ground truth}};
    \end{tikzpicture}
    &
    \begin{tikzpicture}
    \node[anchor=south west,inner sep=0]{\includegraphics[height=3cm]{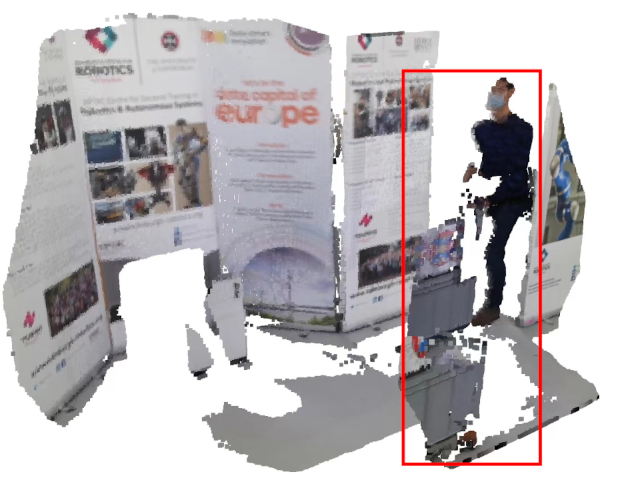}};
    \node[anchor=south west]{\textbf{RF$^*$}};
    \end{tikzpicture}
    \\\hline
    \begin{tikzpicture}
    \node[anchor=south west,inner sep=0]{\includegraphics[height=3cm]{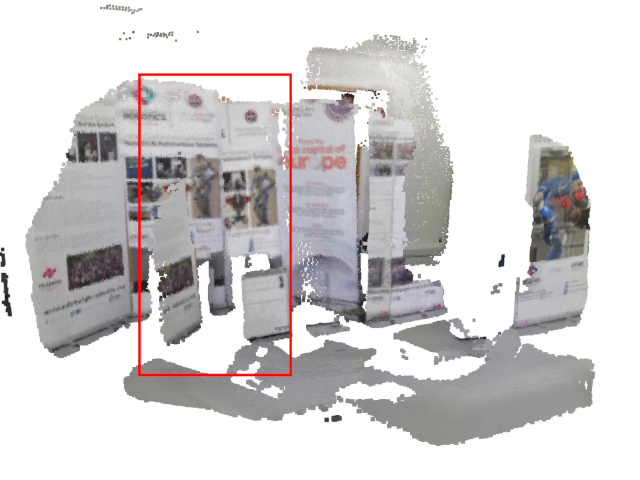}};
    \node[anchor=south west]{\textbf{CF}};
    \end{tikzpicture}
    &
    \begin{tikzpicture}
    \node[anchor=south west,inner sep=0]{\includegraphics[height=3cm]{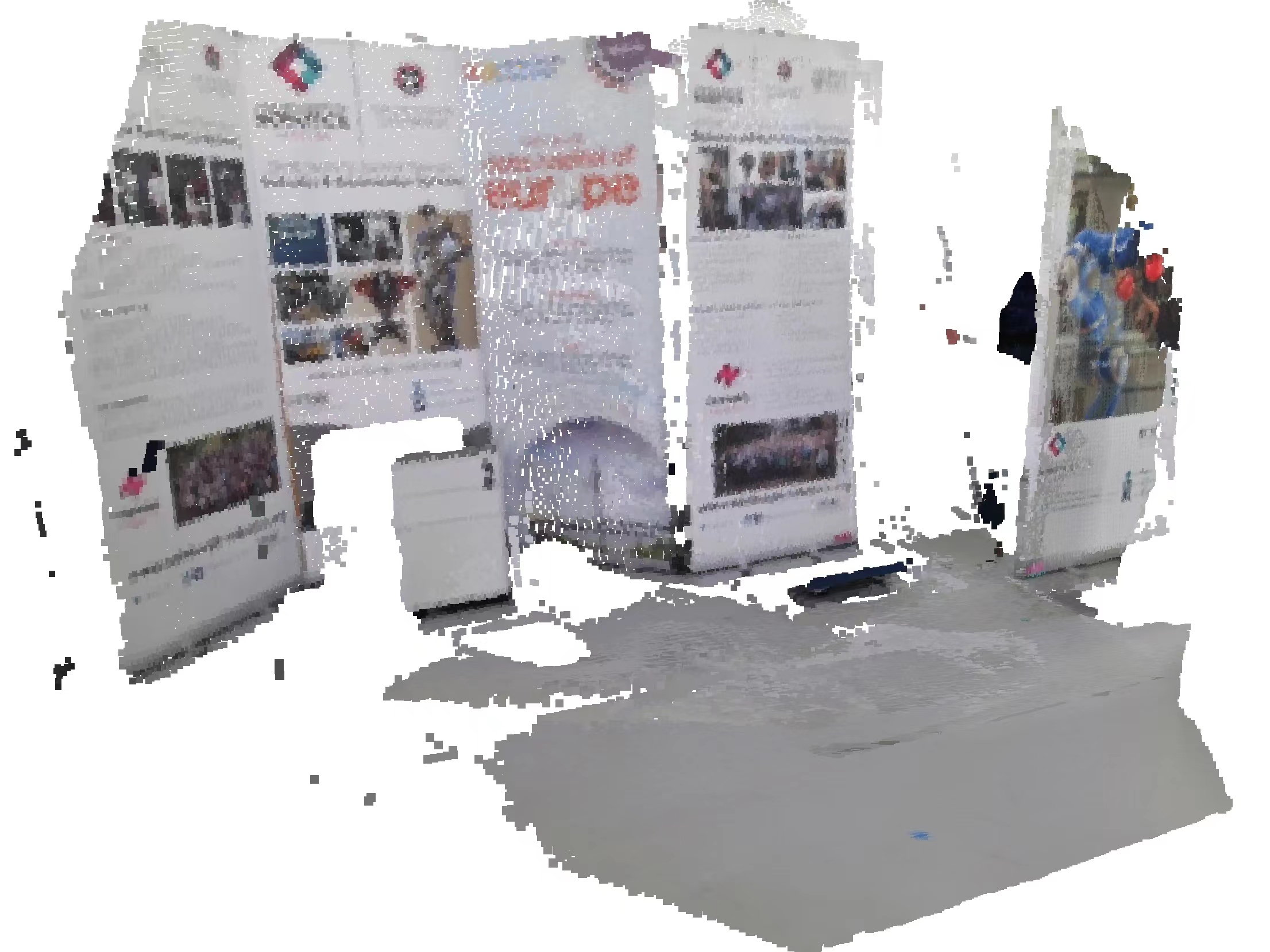}};
    \node[anchor=south west]{\textbf{ours$^*$}};
    \end{tikzpicture}
    \end{tabular}
    \caption{Reconstruction result of the RGB-D sequence 3. The reconstruction failures are marked with red rectangles. RF has mapped dynamic objects into the background. CF has mapped the same static poster twice, which indicates wrong localisation results.}
    \label{fig:reconstruction}
\end{figure}

We qualitatively evaluate the reconstruction result of \textit{seq3} (\Cref{fig:reconstruction}). Since we have no ground truth segmentation, we re-collect a new sequence with the same camera trajectory but no dynamic objects to recover the true background. As shown in the results, RF$^*$ maps the dynamic objects into the static background model. CF has mapped the same static object twice, which is caused by wrong camera pose estimation. Only our proposed method can remove all dynamic objects and correctly reconstruct the background.

\subsection{Planar rigid objects trajectory}
\begin{figure}[t]
    \setlength{\belowcaptionskip}{0cm}
    \centering
    \includegraphics[width=\linewidth]{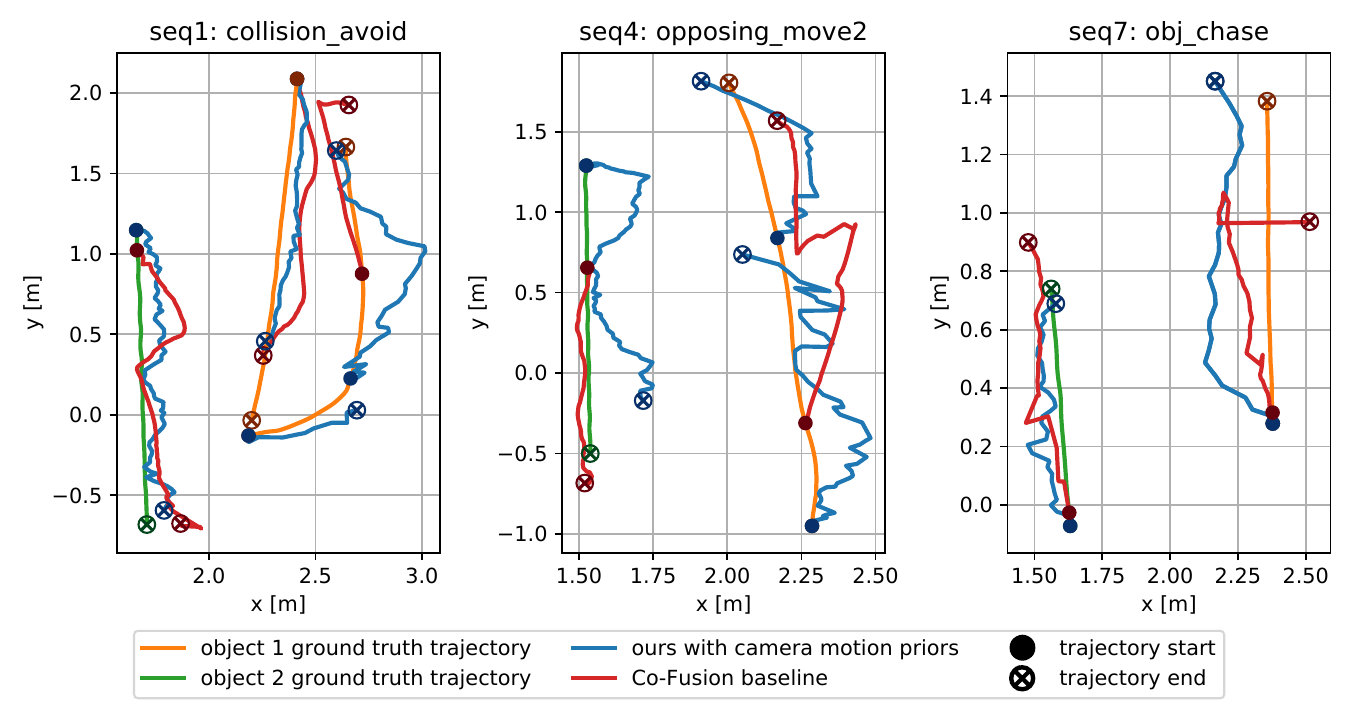}
    \caption{Comparison between CF baseline (red) and our method with the camera motion prior (blue) in terms of the estimated object trajectories. Our method can detect an object as dynamic immediately when it starts to move and provide a more accurate trajectory than CF.
    }
    \label{fig:object_trajectory}
\end{figure}

For both objects, we compute the ATE RMSE between the estimated and ground-truth trajectories when they are in the camera view (\Cref{tab:object_ate}). Since the object can move out of or move into the camera view several times, one object trajectory can be divided into multiple parts. For each object, we, therefore, use the maximal ATE RMSE among the estimated trajectories of different parts for the final result. Our method can provide more accurate and complete object trajectories than CF, but loses track of a dynamic object when the object stops moving or is occluded by other objects (\Cref{fig:object_trajectory}).

\begin{table}[]
\centering
\begin{tabular}{|c|cc|cc|cc|}
\hline
     & \multicolumn{2}{c|}{seq1} & \multicolumn{2}{c|}{seq4} & \multicolumn{2}{c|}{seq7} \\ 
     \cline{2-7}
     & \multicolumn{1}{c|}{object1}& object2& \multicolumn{1}{c|}{object1}& object2& \multicolumn{1}{c|}{object1}& object2\\ \hline
CF   & \multicolumn{1}{c}{21.5}   & 10.6   & \multicolumn{1}{c}{24.2}   & \textbf{5.36} & \multicolumn{1}{c}{33.8}   & 6.57   \\ \hline
CF$^{*}$   & \multicolumn{1}{c}{20.9}   & 16.3   & \multicolumn{1}{c}{20.5}   & 6.21 & \multicolumn{1}{c}{17.1}   & 12.9   \\ \hline
ours$^{*}$ & \multicolumn{1}{c}{\textbf{13.1}} & \textbf{4.95} & \multicolumn{1}{c}{\textbf{4.95}} & 8.84   & \multicolumn{1}{c}{\textbf{7.27}} & \textbf{3.93} \\ \hline
\end{tabular}
\caption{ATE RMSE of the object trajectories estimated from CF baseline, CF$^{*}$ and ours$^{*}$.}
\label{tab:object_ate}
\end{table}

\subsection{Impact of drift in motion prior}
\begin{figure}[t]
    \centering
    \includegraphics[width=\linewidth]{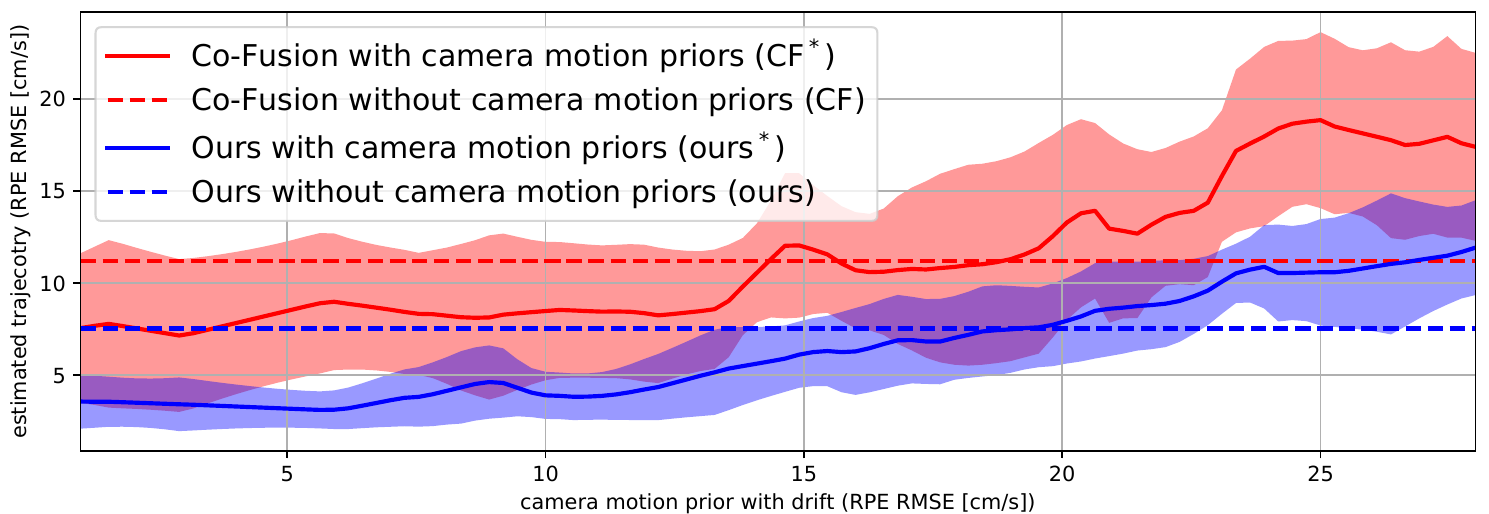}
    \caption{RPE RMSE of the estimated trajectories against the drift magnitude of wheel odometry. Our method preforms better than CF when using the camera motion prior with the same magnitude of drift and is robust to nearly 24 cm/s odometry drift until it is comparable with CF baseline.}
    \label{fig:rpe_drift}
\end{figure}

We increase the drift magnitude of the camera motion prior to test our methods' robustness to different levels of drift. By comparing the RPE RMSE of the camera motion prior and estimated trajectories, we find that our method can outperform Co-Fusion baseline with drift up to 24 cm/s (\Cref{fig:rpe_drift}). Even when the motion prior has a drift of nearly 30 cm/s, we can still reduce the drift to around 12 cm/s. Compared to Co-Fusion with camera motion prior, our method is always better using the motion prior with the same magnitude of drift. 

\section{Conclusion}
\label{sec:conclusion}
This work presented a dense RGB-D SLAM method that tracks multiple planar rigid objects without relying on semantic segmentation. We also proposed a novel online multimotion segmentation method and a dynamic segmentation pipeline based on a hierarchical representation of planes and super-pixels. The detailed evaluation demonstrates that our method achieves better localisation and mapping results than state-of-the-art approaches when multiple dynamic objects occupy the major proportion of the camera view. If one dynamic object is occluded by another, our method fails to track the object but detects the object as new after it reappears in the camera view. Our future work would be re-detecting the dynamic objects based on their models to support long-term object tracking. We also plan to extend our method to non-planar environments and enable independently tracking of multiple large non-planar rigid objects.


\bibliographystyle{IEEEtran.bst}
\bibliography{IEEEabrv,mybibfile}
\end{document}